\documentclass[letterpaper]{article} 
\usepackage{aaai24}  
\usepackage{times}  
\usepackage{helvet}  
\usepackage{courier}  
\usepackage[hyphens]{url}  
\usepackage{graphicx} 
\usepackage{natbib}  
\usepackage{caption} 
\usepackage{algorithm}
\usepackage{algorithmic}

\usepackage{tabularx}
\usepackage{amsmath}
\usepackage{amssymb}
\usepackage{graphicx}
\usepackage{booktabs}  
\usepackage{threeparttable} \usepackage{multirow} 

\usepackage{subfigure}

\newtheorem{theorem}{Theorem}

\newtheorem{prop}{Proposition}

\newtheorem{definition}{Definition}
\newtheorem{assumption}{Assumption}

\usepackage{newfloat}
\usepackage{listings}
\DeclareCaptionStyle{ruled}{labelfont=normalfont,labelsep=colon,strut=off} 
\floatstyle{ruled}
\newfloat{listing}{tb}{lst}{}
\floatname{listing}{Listing}
\title{Provably Convergent Federated Trilevel Learning}
\author {
    Yang Jiao\textsuperscript{\rm 1},
    Kai Yang\textsuperscript{\rm 1,2,3}\thanks{Corresponding author (e-mail: kaiyang@tongji.edu.cn).},
    Tiancheng Wu\textsuperscript{\rm 1},
    Chengtao Jian\textsuperscript{\rm 1},
    Jianwei Huang\textsuperscript{\rm 4,5}
}
\affiliations {
    \textsuperscript{\rm 1}Department of Computer Science and Technology, Tongji University\\
    \textsuperscript{\rm 2} Key Laboratory of Embedded System and Service Computing Ministry of Education at Tongji University\\
    \textsuperscript{\rm 3}Shanghai Research Institute for Intelligent Autonomous Systems \\
    \textsuperscript{\rm 4}School of Science and Engineering, The Chinese University of Hong Kong, Shenzhen\\
    \textsuperscript{\rm 5}Shenzhen Institute of Artificial Intelligence and Robotics for Society\\
    
    yangjiao@tongji.edu.cn, kaiyang@tongji.edu.cn, tony318@tongji.edu.cn, jct@tongji.edu.cn, \\
    jianweihuang@cuhk.edu.cn
}

\usepackage{bibentry}

\begin{document}

\maketitle

\begin{abstract}
Trilevel learning, also called trilevel optimization (TLO), has been recognized as a powerful modelling tool for hierarchical decision process and widely applied in many machine learning applications, such as robust neural architecture search, hyperparameter optimization, and domain adaptation. Tackling TLO problems has presented a great challenge due to their nested decision-making structure. In addition, existing works on TLO face the following key challenges: 1) they all focus on the non-distributed setting, which may lead to privacy breach; 2) they do not offer any non-asymptotic convergence analysis which characterizes how fast an algorithm converges. To address the aforementioned challenges, this paper proposes an asynchronous federated trilevel optimization method to solve TLO problems. The proposed method utilizes $\mu$-cuts to construct a hyper-polyhedral approximation for the TLO problem and solve it in an asynchronous manner. We demonstrate that the proposed $\mu$-cuts are applicable to not only convex functions but also a wide range of non-convex functions that meet the $\mu$-weakly convex assumption. Furthermore, we theoretically analyze the non-asymptotic convergence rate for the proposed method by showing its iteration complexity to obtain $\epsilon$-stationary point is upper bounded by $\mathcal{O}(\frac{1}{\epsilon^2})$. Extensive experiments on real-world datasets have been conducted to elucidate the superiority of the proposed method, e.g., it has a  faster convergence rate with a maximum acceleration of approximately 80$\%$.

\end{abstract}

\section{Introduction}

Recently, trilevel learning, also called trilevel optimization (TLO), has found applications in many machine learning tasks, e.g., robust neural architecture search \cite{guo2020meets}, robust hyperparameter optimization \cite{sato2021gradient} and domain adaptation \cite{raghu2021meta}. Trilevel optimization problems refer to the optimization problems that involve three-level optimization problems and thus have a trilevel hierarchy \cite{avraamidou2018mixed,sato2021gradient}. A general form of trilevel optimization problem is given by,
\begin{equation}
\label{eq:trilevel}
\begin{array}{l}
\min  {{f_{1}}({\boldsymbol{x}_1},{\boldsymbol{x}_2},{\boldsymbol{x}_3})} \;{\rm{s}}.{\rm{t}}. \vspace{0.5mm}\\
\qquad {\boldsymbol{x}_2} = \mathop {\arg \min }\limits_{{\boldsymbol{x}_2}'}  {{f_{2}}({\boldsymbol{x}_1},{\boldsymbol{x}_2}',{\boldsymbol{x}_3})} \;{\rm{s}}.{\rm{t}}. \\
\qquad \qquad {\boldsymbol{x}_3} = \mathop {\arg \min }\limits_{{\boldsymbol{x}_3}'}  {{f_{3}}({\boldsymbol{x}_1},{\boldsymbol{x}_2}',{\boldsymbol{x}_3}')} \\
{\mathop{\rm var}}. \qquad \qquad {\boldsymbol{x}_1},{\boldsymbol{x}_2},{\boldsymbol{x}_3},
\end{array}
\end{equation}
where $f_1, f_2, f_3$ respectively denote the first, second, and third level objectives. Here $\boldsymbol{x}_1 \!\in\! \mathbb{R}^{d_1}, \boldsymbol{x}_2 \!\in\! \mathbb{R}^{d_2}, \boldsymbol{x}_3 \!\in\! \mathbb{R}^{d_3}$ are variables. Despite its wide applications, the development of solution methods was predominately limited
to bilevel optimization (BLO) \cite{ji2021bilevel,franceschi2018bilevel} primarily due to the escalated difficulty in solving the TLO problem \cite{sato2021gradient}. The literature, specifically \cite{blair1992computational,avraamidou2018mixed}, highlights that the complexity associated with solving problems characterized by hierarchical structures comprising more than two levels is substantially greater compared to that of bilevel optimization problems.

Theoretical work on solving TLO problems only emerge during the recent several years. A hypergradient (gradient)-based method is proposed in \cite{sato2021gradient}, which uses $K$ gradient descent steps to replace the lower-level problem to solve the TLO problems. This algorithm in \cite{sato2021gradient} is one of the first results that establish theoretical guarantees for solving the TLO problem.  A general automatic differentiation technique is proposed in \cite{choe2022betty}, which is based on the interpretation of TLO as a special type of dataflow graph.
Nevertheless, there are still some issues that have not been addressed in the prior work, including 1) in TLO applications, data may be acquired and disseminated across multiple nodes, the prior works only solve the TLO problems in a non-distributed manner, which needs to collect a massive amount of data to a single server and may lead to the data privacy risks  \cite{subramanya2021centralized,jiao2022asynchronous,han2020adaptive}. Moreover, the synchronous federated algorithms often suffer from straggler
problems and will immediately stop working if some workers fail to communicate \cite{jiao2022asynchronous}. Therefore, developing asynchronous federated algorithms for TLO is significantly important. 2) The existing TLO works only provide the asymptotic convergence guarantee for their algorithms. In order to understand the convergence speed of the proposed algorithm, non-asymptotic convergence analysis that can characterize how fast an algorithm converges in practice is required. 

To this end, we propose an \textbf{A}synchronous \textbf{F}ederated \textbf{T}rilevel \textbf{O}ptimization method (AFTO) in this paper. The proposed AFTO can effectively solve the TLO problems in an asynchronous federated manner. Specifically, it treats the lower-level optimization problem as a constraint to the upper-level and utilizes $\mu$-cuts to construct the hyper-polyhedral approximation, then an effective asynchronous algorithm is developed.  In the context of trilevel learning problems, the objective functions at each level are usually non-convex, thus the cutting plane methods tailored for convex functions \cite{jiao2022asynchronous,franc2011cutting} are found to be inapplicable. To our best knowledge, the proposed methodology referred to as  $\mu$-cut represents the first approach that is capable of constructing cutting planes for trilevel learning problems characterized by non-convex objectives. Furthermore, we demonstrate that the proposed method is guaranteed to converge and theoretically analyze the non-asymptotic convergence rate in terms of iteration complexity.  


The contributions of this work are summarized as follows.

\textbf{1.} An asynchronous federated trilevel optimization method is proposed in this work for trilevel learning. To our best knowledge, it is the first work designing algorithms to solve the trilevel learning problem in an asynchronous distributed manner.

\textbf{2.} A novel hyper-polyhedral approximation method via  $\mu$-cut is proposed in this work. The proposed $\mu$-cut can be applied to trilevel learning with non-convex objectives. We further demonstrate that the iteration complexity of the proposed method to achieve the $\epsilon$-stationary point is upper bounded by $\mathcal{O}(\frac{1}{\epsilon^2})$.

\textbf{3.} Extensive experiments on real-world datasets justify the superiority of the proposed method and underscore the significant benefits of employing the hyper-polyhedral approximation for trilevel learning. 

\section{Related Work}
\subsection{Trilevel Optimization}
Trilevel optimization has many applications ranging from economics to machine learning. A robust neural architecture search approach is proposed in \cite{guo2020meets}, which integrates the adversarial training into one-shot neural architecture search and can be regarded as solving a trilevel optimization problem. TimeAutoAD \cite{jiao2022timeautoad} is proposed to automatically configure the anomaly detection pipeline and optimize the hyperparameters for multivariate time series anomaly detection. The optimization problem that TimeAutoAD aims to solve can be viewed as a trilevel optimization problem. And a method is proposed in \cite{raghu2021meta} to solve the trilevel optimization problem which involves hyperparameter optimization and two-level pretraining and finetuning. LFM \cite{garg2022learning} is proposed to solve a trilevel optimization problem which consists of data reweight, architecture search, and model training. A general automatic differentiation technique Betty is proposed in \cite{choe2022betty}, which can be utilized to solve the trilevel optimization problem. However, the aforementioned algorithms do not provide any convergence guarantee. A hypergradient-based algorithm with asymptotic convergence guarantee is proposed in \cite{sato2021gradient}, which can be employed in trilevel optimization problems. Nevertheless, the existing works focus on solving the TLO problems in a non-distributed manner and do not provide any non-asymptotic convergence analysis. Instead, an efficient asynchronous algorithm with non-asymptotic convergence guarantee is proposed in this work for solving TLO problems. To our best knowledge, this is the first work that solves TLO problems in an asynchronous federated manner.

\subsection{Polyhedral Approximation}
Polyhedral approximation is a widely-used approximation method \cite{bertsekas2015convex}. The idea behind polyhedral approximation is to approximate either the feasible region or the 
epigraph of the objective function of an optimization problem by a set of cutting planes, and the approximation will be gradually refined by adding additional cutting planes. Since the approximate problem is polyhedral, it is usually much easier to solve than the original problem. Following \cite{bertsekas2015convex}, the polyhedral approximation can be broadly divided into two main approaches: outer linearization and inner linearization. The outer linearization \cite{tawarmalani2005polyhedral,yang2008distributed,burger2013polyhedral} (also called cutting plane method) utilizes a set of cutting planes to approximate the feasible region or the 
epigraph of the objective function from 
without. In contrast, inner linearization \cite{bertsekas2011unifying,trombettoni2011inner} utilizes the convex hulls 
of finite numbers of halflines or points to approximate the feasible region or the
epigraph of the objective function from within. Polyhedral approximation has been widely used in convex optimization. A polyhedral approximation method is proposed in \cite{bertsekas2015convex} for convex optimization, which utilizes cutting planes to approximate the original convex optimization problem. In \cite{burger2013polyhedral}, a fully distributed
algorithm is proposed, which is based on an outer polyhedral approximation of
the constraint sets, for the convex and robust distributed optimization
problems in peer-to-peer networks.  In this work, a novel hyper-polyhedral approximation method via $\mu$-cut is proposed for TLO. The proposed $\mu$-cut can be utilized for $\mu$-weakly convex optimization and thus has broader applicability compared with the cutting plane methods for convex optimization.

\section{Asynchronous Federated Trilevel Learning}
Traditional trilevel optimization methods require collecting a massive amount of data to a single server for model training, which may lead to data privacy risks. Solving trilevel optimization problems in a distributed manner is challenging since the trilevel optimization problem is highly-nested which hinders the development of the distributed algorithms.  
The distributed trilevel optimization problem can be expressed as,
\begin{equation}
\label{eq:1}
\begin{array}{l}
\min \sum\nolimits_{j = 1}^N {{f_{1,j}}({\boldsymbol{x}_1},{\boldsymbol{x}_2},{\boldsymbol{x}_3})} \;{\rm{s}}.{\rm{t}}. \vspace{0.25mm}\\
\qquad {\boldsymbol{x}_2} = \mathop {\arg \min }\limits_{{\boldsymbol{x}_2}'} \sum\nolimits_{j = 1}^N {{f_{2,j}}({\boldsymbol{x}_1},{\boldsymbol{x}_2}',{\boldsymbol{x}_3})} \;{\rm{s}}.{\rm{t}}. \\
\qquad \qquad {\boldsymbol{x}_3} = \mathop {\arg \min }\limits_{{\boldsymbol{x}_3}'} \sum\nolimits_{j = 1}^N {{f_{3,j}}({\boldsymbol{x}_1},{\boldsymbol{x}_2}',{\boldsymbol{x}_3}')} \vspace{0.5mm}\\
{\mathop{\rm var}}. \qquad \qquad {\boldsymbol{x}_1},{\boldsymbol{x}_2},{\boldsymbol{x}_3},
\end{array}
\end{equation}
where $N$ denotes the number of workers in distributed systems, $f_{1,j}, f_{2,j}, f_{3,j}$ denote the local first, second, and third level objectives in worker $j$, respectively. The problem in Eq. (\ref{eq:1}) can be reformulated as a consensus problem \cite{zhang2014asynchronous,jiao2022distributed},
\begin{equation}
\label{eq:2}
\begin{array}{l}
\min \sum\nolimits_{j} {{f_{1,j}}({\boldsymbol{x}_{1,j}},{\boldsymbol{x}_{2,j}},{\boldsymbol{x}_{3,j}})} \;{\rm{s}}.{\rm{t}}. \\

 {\boldsymbol{x}_{1,j}} = {\boldsymbol{z}_1},j = 1, \cdots ,N \\
 
\! \{{\boldsymbol{x}_{2,j}}\} ,{\boldsymbol{z}_2} \!=\! \mathop {\arg \min }\limits_{\{ {\boldsymbol{x}_{2,j}}'\} ,{\boldsymbol{z}_2}'} \sum\nolimits_{j} {{f_{2,j}}({\boldsymbol{z}_1},{\boldsymbol{x}_{2,j}}',{\boldsymbol{x}_{3,j}})} \;{\rm{s}}.{\rm{t}}. \\

 \qquad \quad \;\;\, \,  {\boldsymbol{x}_{2,j}}' = {\boldsymbol{z}_2}',j = 1, \cdots ,N \\

 \qquad \quad \;\;\,   \{{\boldsymbol{x}_{3,j}}\} ,{\boldsymbol{z}_3} \!=\! \mathop {\arg \min }\limits_{\{{\boldsymbol{x}_{3,j}}'\},{\boldsymbol{z}_3}'} \! \sum\nolimits_{j} \!{{f_{3,j}}({\boldsymbol{z}_1},{\boldsymbol{z}_2}',{\boldsymbol{x}_{3,j}}')} \,{\rm{s}}.{\rm{t}}. \\

 \qquad \qquad \qquad \qquad \quad   {\boldsymbol{x}_{3,j}}' = {\boldsymbol{z}_3}',j = 1, \cdots ,N \vspace{0.5mm}\\

{\mathop{\rm var}}. \qquad  \qquad \{{\boldsymbol{x}_{1,j}}\} ,{\rm{\{ }}{\boldsymbol{x}_{2,j}}\} ,{\rm{\{ }}{\boldsymbol{x}_{3,j}}\} ,{\boldsymbol{z}_1}{\rm{,}}{\boldsymbol{z}_2},{\boldsymbol{z}_3},
\end{array}
\end{equation}
where $\boldsymbol{x}_{1,j} \!\in\! \mathbb{R}^{d_1}, \boldsymbol{x}_{2,j}\!\in\! \mathbb{R}^{d_2}, \boldsymbol{x}_{3,j}\!\in\! \mathbb{R}^{d_3}$ denote the local variables in worker $j$, and $\boldsymbol{z}_1 \!\in\! \mathbb{R}^{d_1}, \boldsymbol{z}_2\!\in\! \mathbb{R}^{d_2}, \boldsymbol{z}_3\!\in\! \mathbb{R}^{d_3}$ denote the consensus variables in the master. This reformulation in Eq. (\ref{eq:2}) can facilitate the development of distributed algorithms for trilevel optimization problems based on the parameter-server architecture \cite{assran2020advances}. The remaining procedure of the proposed method can be divided into three steps. First, how to construct the hyper-polyhedral approximation for distributed TLO problems is proposed. Then, an effective asynchronous federated algorithm is developed. Finally, how to update the $\mu$-cuts to refine the hyper-polyhedral approximation is proposed.

\subsection{Hyper-Polyhedral Approximation}
Different from the traditional polyhedral approximation method \cite{bertsekas2015convex,franc2011cutting,burger2013polyhedral}, a novel hyper-polyhedral approximation method is proposed for distributed TLO problems in this work. By utilizing the proposed hyper-polyhedral approximation, 
the distributed algorithms can be easier to develop for TLO problems. Specifically, the proposed hyper-polytope consists of the ${\rm{\uppercase\expandafter{\romannumeral1}}}^{\rm{st}}$ layer and ${\rm{\uppercase\expandafter{\romannumeral2}}}^{\rm{nd}}$  layer polytopes, which are introduced as follows.

\subsubsection{${\rm{\uppercase\expandafter{\romannumeral1}}}^{\rm{st}}$ layer Polyhedral Approximation:}

First, defining  $h_{\rm{\uppercase\expandafter{\romannumeral1}}}(\{{\boldsymbol{x}_{3,j}}\},\boldsymbol{z}_1, {\boldsymbol{z}_2}',\boldsymbol{z}_3 ) \!=\!||\left[ \begin{array}{l}
    \! \{{\boldsymbol{x}_{3,j}}\} \! \\
     \boldsymbol{z}_3 
\end{array} \right]  \!-\!\phi_{{\rm{\uppercase\expandafter{\romannumeral1}}}}({\boldsymbol{z}_1},{\boldsymbol{z}_2}')||^2 $ and $\phi_{\rm{\uppercase\expandafter{\romannumeral1}}}({\boldsymbol{z}_1},{\boldsymbol{z}_2}')\!=\! \mathop{\arg \min }\nolimits_{\{{\boldsymbol{x}_{3,j}}'\},{\boldsymbol{z}_3}'}\{ \sum\nolimits_{j=1}^N \! {{f_{3,j}}({\boldsymbol{z}_1},{\boldsymbol{z}_2}',{\boldsymbol{x}_{3,j}}')}: {\boldsymbol{x}_{3,j}}' \!=\! {\boldsymbol{z}_3}',\forall j\}$.  In trilevel optimization, the third level optimization problem can be viewed as the constraint to the second level optimization problem \cite{chen2022trilevel}, i.e., $h_{\rm{\uppercase\expandafter{\romannumeral1}}}(\{{\boldsymbol{x}_{3,j}}\},\boldsymbol{z}_1, {\boldsymbol{z}_2}',\boldsymbol{z}_3 )\!=\!0$.  A consensus problem needs to be solved in a distributed manner if the exact $\phi_{\rm{\uppercase\expandafter{\romannumeral1}}}({\boldsymbol{z}_1},{\boldsymbol{z}_2}')$ is required. In many works in bilevel \cite{ji2021bilevel,liu2021towards,jiao2022asynchronous} and trilevel \cite{sato2021gradient} optimization, the exact $\phi_{\rm{\uppercase\expandafter{\romannumeral1}}}({\boldsymbol{z}_1},{\boldsymbol{z}_2}')$ can be replaced by an estimate of $\phi_{\rm{\uppercase\expandafter{\romannumeral1}}}({\boldsymbol{z}_1},{\boldsymbol{z}_2}')$, and we utilize the results after $K$ communication rounds between the master and workers as the estimate of $\phi_{\rm{\uppercase\expandafter{\romannumeral1}}}({\boldsymbol{z}_1},{\boldsymbol{z}_2}')$ according to \cite{jiao2022asynchronous}. Specifically, for the third level optimization problem, the augmented Lagrangian function can be written as,
\begin{equation}
\begin{array}{l}
\! L_{p,3} \!=\! \sum\nolimits_{j = 1}^N ( {{f_{3,j}}({\boldsymbol{z}_1},{\boldsymbol{z}_2}'\!,{\boldsymbol{x}_{3,j}}')} + {\boldsymbol{\varphi}_{3,j}^{\top}}({\boldsymbol{x}_{3,j}}'- {\boldsymbol{z}_3}') \vspace{0.5mm}\\ 
\qquad \qquad \quad \, + \frac{\kappa_3}{2}||{\boldsymbol{x}_{3,j}}'- {\boldsymbol{z}_3}'||^2 ),
\end{array}
\end{equation}
where $L_{p,3} = L_{p,3}({\boldsymbol{z}_1},\!{\boldsymbol{z}_2}'\!, {\boldsymbol{z}_3}'\!,\! \{{\boldsymbol{x}_{3,j}}'\},\! \{\boldsymbol{\varphi}_{3,j}\})$, $\boldsymbol{\varphi}_{3,j} \!\in\! \mathbb{R}^{d_3}$ is the dual variable, and constant $\kappa_3\!>\!0$ is a penalty parameter. In $(k+1)^{\rm{th}}$ communication round, we have that,

\noindent 1) Workers update the local variables,
\begin{equation}
{\boldsymbol{x}_{3,j}^{k+1}}' \!=\! {\boldsymbol{x}_{3,j}^{k}}'\!-\!\eta_{\boldsymbol{x}} \!\nabla_{\boldsymbol{x}_{3,j}} L_{p,3}({\boldsymbol{z}_1},{\boldsymbol{z}_2}'\!, {\boldsymbol{z}_3^k}',\! \{{\boldsymbol{x}_{3,j}^k}'\},\! \{\boldsymbol{\varphi}_{3,j}^k\}),
\end{equation}
where $\eta_{\boldsymbol{x}}$ represents the step-size. Then, workers transmit the local variables ${\boldsymbol{x}_{3,j}^{k+1}}'$ to the master.

\noindent 2) Master updates the variables as follows,
\begin{equation}
{\boldsymbol{z}_3^{k+1}}' \!=\! {\boldsymbol{z}_3^k}' \!-\! \eta_{\boldsymbol{z}}\! \nabla_{\boldsymbol{z}_3} L_{p,3}({\boldsymbol{z}_1},{\boldsymbol{z}_2}' \!, {\boldsymbol{z}_3^k}', \! \{{\boldsymbol{x}_{3,j}^k}'\}, \! \{\boldsymbol{\varphi}_{3,j}^k\}),
\end{equation}
\begin{equation}
\boldsymbol{\varphi}_{3,j}^{k+1} \!=\! \boldsymbol{\varphi}_{3,j}^{k}+\eta_{\boldsymbol{\varphi}}\! \nabla_{\boldsymbol{\varphi}_{3,j}} L_{p,3}({\boldsymbol{z}_1},{\boldsymbol{z}_2}' \!, {\boldsymbol{z}_3^{k+1}}', \! \{{\boldsymbol{x}_{3,j}^{k+1}}'\}, \! \{\boldsymbol{\varphi}_{3,j}^k\}),
\end{equation}
where $\eta_{\boldsymbol{z}}$ and $\eta_{\boldsymbol{\varphi}}$ represent the step-sizes. Then, master broadcasts the ${\boldsymbol{z}_3^{k+1}}'$ and $\boldsymbol{\varphi}_{3,j}^{k+1}$ to workers.

The results after $K$ communication rounds are utilized as the estimate of $\phi_{\rm{\uppercase\expandafter{\romannumeral1}}}({\boldsymbol{z}_1},{\boldsymbol{z}_2}')$, that is,
\begin{equation}
\begin{array}{l}
\label{4_17_12}
\phi_{\rm{\uppercase\expandafter{\romannumeral1}}}({\boldsymbol{z}_1},{\boldsymbol{z}_2}') 
\! = \left[ \begin{array}{l}
  \! \{ {\boldsymbol{x}_{3,j}^{0}}' \!-\!\sum\nolimits_{k = 0}^{K-1} \! \eta_{\boldsymbol{x}} \! \nabla_{\boldsymbol{x}_{3,j}} L_{p,3}^k \} \! \vspace{0.5mm} \\
  
  {\boldsymbol{z}_3^{0}}'-\sum\nolimits_{k = 0}^{K-1}\eta_{\boldsymbol{z}} \nabla_{\boldsymbol{z}_3} L_{p,3}^k
\end{array} \right] ,
\end{array}
\end{equation}
where $L_{p,3}^k = L_{p,3}({\boldsymbol{z}_1},{\boldsymbol{z}_2}', {\boldsymbol{z}_3^k}', \{{\boldsymbol{x}_{3,j}^k}'\}, \{\boldsymbol{\varphi}_{3,j}^k\})$. Based on Eq. (\ref{4_17_12}) and the definition of $h_{\rm{\uppercase\expandafter{\romannumeral1}}}$, we have that,
\begin{equation} 
\begin{array}{l}
\label{eq:5_8_9}
\!h_{\rm{\uppercase\expandafter{\romannumeral1}}}(\{{\boldsymbol{x}_{3,j}}\},\boldsymbol{z}_1, {\boldsymbol{z}_2}',\boldsymbol{z}_3 ) \\ \!=\! ||\! \left[ \! \begin{array}{l}
 \! \{ \boldsymbol{x}_{3,j} - {\boldsymbol{x}_{3,j}^{0}}'+\sum\nolimits_{k = 0}^{K-1} \!
 \eta_{\boldsymbol{x}} \nabla_{\boldsymbol{x}_{3,j}} L_{p,3}^k \} \!  \vspace{0.5mm}\\
  
  \boldsymbol{z}_3 - {\boldsymbol{z}_3^{0}}'+\sum\nolimits_{k = 0}^{K-1}\eta_{\boldsymbol{z}} \nabla_{\boldsymbol{z}_3} L_{p,3}^k
\end{array} \! \right] \! ||^2 .
\end{array}
\end{equation}

Inspired by polyhedral approximation method \cite{bertsekas2015convex,burger2013polyhedral}, the \textbf{${\rm{\uppercase\expandafter{\romannumeral1}}}^{\rm{st}}$ layer} \textbf{polytope}, which forms of a set of cutting planes (i.e., linear inequalities), is utilized to approximate the feasible region with respect to the constraint $h_{\rm{\uppercase\expandafter{\romannumeral1}}}(\{{\boldsymbol{x}_{3,j}}\},\boldsymbol{z}_1, {\boldsymbol{z}_2}',\boldsymbol{z}_3 )\! \le \! \varepsilon_{\rm{\uppercase\expandafter{\romannumeral1}}}$, which is a relaxed form of constraint $h_{\rm{\uppercase\expandafter{\romannumeral1}}}(\{{\boldsymbol{x}_{3,j}}\},\boldsymbol{z}_1, {\boldsymbol{z}_2}',\boldsymbol{z}_3 )\!=\!0$ in Eq. (\ref{eq:5_8_9}), and $\varepsilon_{\rm{\uppercase\expandafter{\romannumeral1}}}\!>\!0$  is a pre-set constant. Specifically, the ${\rm{\uppercase\expandafter{\romannumeral1}}}^{\rm{st}}$ layer polytope in $(t+1)^{\rm{th}}$ iteration can be expressed as $P_{\rm{\uppercase\expandafter{\romannumeral1}}}^{t}\!=\!\{{\boldsymbol{a}_{1,l}^{\rm{\uppercase\expandafter{\romannumeral1}}}}^{\top}\!{\boldsymbol{z}_1} + {\boldsymbol{a}_{2,l}^{\rm{\uppercase\expandafter{\romannumeral1}}}}^{\top}\!{\boldsymbol{z}_2}' + {\boldsymbol{a}_{3,l}^{\rm{\uppercase\expandafter{\romannumeral1}}}}^{\top}\!{\boldsymbol{z}_3} + \sum\nolimits_{j = 1}^N {\boldsymbol{b}{{_{j,l}^{\rm{\uppercase\expandafter{\romannumeral1}}}}}}^{\top}\!{\boldsymbol{x}_{3,j}}  \!\le\! c_l^{\rm{\uppercase\expandafter{\romannumeral1}}},l \!=\! 1,\! \cdots\! ,|P_{\rm{\uppercase\expandafter{\romannumeral1}}}^{t}| \}$, where $|P_{\rm{\uppercase\expandafter{\romannumeral1}}}^{t}|$ denotes the number of cutting planes in ${\rm{\uppercase\expandafter{\romannumeral1}}}^{\rm{st}}$ layer polytope and ${\boldsymbol{a}_{i,l}^{\rm{\uppercase\expandafter{\romannumeral1}}}}, {\boldsymbol{b}_{j,l}^{\rm{\uppercase\expandafter{\romannumeral1}}}}, {c_{l}^{\rm{\uppercase\expandafter{\romannumeral1}}}}$ are parameters in $l^{\rm{th}}$ cutting plane (${\rm{\uppercase\expandafter{\romannumeral1}}}^{\rm{st}}$ layer $\mu$-cut). Defining $\hat{h}_{{\rm{\uppercase\expandafter{\romannumeral1}}},l}(\{{\boldsymbol{x}_{3,j}}\},\boldsymbol{z}_1, {\boldsymbol{z}_2}',\boldsymbol{z}_3 ) = {\boldsymbol{a}_{1,l}^{\rm{\uppercase\expandafter{\romannumeral1}}}}^{\top}\!{\boldsymbol{z}_1} + {\boldsymbol{a}_{2,l}^{\rm{\uppercase\expandafter{\romannumeral1}}}}^{\top}\!{\boldsymbol{z}_2}' + {\boldsymbol{a}_{3,l}^{\rm{\uppercase\expandafter{\romannumeral1}}}}^{\top}\!{\boldsymbol{z}_3} + \sum\nolimits_{j = 1}^N {\boldsymbol{b}{{_{j,l}^{\rm{\uppercase\expandafter{\romannumeral1}}}}}}^{\top}\!{\boldsymbol{x}_{3,j}} $, the resulting (bilevel) problem can be expressed as,
\begin{equation}
\label{eq:4}
 \begin{array}{l}
\min \sum\nolimits_{j = 1}^N {{f_{1,j}}({\boldsymbol{x}_{1,j}},{\boldsymbol{x}_{2,j}},{\boldsymbol{x}_{3,j}})} \;{\rm{s}}.{\rm{t}}. \\

 {\boldsymbol{x}_{1,j}} = {\boldsymbol{z}_1},j = 1, \cdots ,N \\

 \{ {\boldsymbol{x}_{2,j}}\} ,{\boldsymbol{z}_2} \!=\! \mathop {\arg \min }\limits_{\{ {\boldsymbol{x}_{2,j}}'\} ,{\boldsymbol{z}_2}'} \sum\nolimits_{j = 1}^N {{f_{2,j}}({\boldsymbol{z}_1},{\boldsymbol{x}_{2,j}}',{\boldsymbol{x}_{3,j}})} \;{\rm{s}}.{\rm{t}}. \\

\qquad \qquad \quad   {\boldsymbol{x}_{2,j}}' = {\boldsymbol{z}_2}',j = 1, \cdots ,N \\

\qquad \qquad \quad   \hat{h}_{{\rm{\uppercase\expandafter{\romannumeral1}}},l}(\{{\boldsymbol{x}_{3,j}}\},\boldsymbol{z}_1, {\boldsymbol{z}_2}',\boldsymbol{z}_3 )  \!\le\! c_l^{\rm{\uppercase\expandafter{\romannumeral1}}},l \!=\! 1,\! \cdots\! ,|P_{\rm{\uppercase\expandafter{\romannumeral1}}}^{t}|\vspace{0.5mm}\\

{\mathop{\rm var}}. \qquad  \quad \{{\boldsymbol{x}_{1,j}}\} ,{\rm{\{ }}{\boldsymbol{x}_{2,j}}\} ,{\rm{\{ }}{\boldsymbol{x}_{3,j}}\} ,{\boldsymbol{z}_1}{\rm{,}}{\boldsymbol{z}_2},{\boldsymbol{z}_3}.
\end{array}
\end{equation}

\subsubsection{${\rm{\uppercase\expandafter{\romannumeral2}}}^{\rm{nd}}$ layer Polyhedral Approximation:}
Defining function $h_{{\rm{\uppercase\expandafter{\romannumeral2}}}}(\{{\boldsymbol{x}_{2,j}}\} ,\{ {\boldsymbol{x}_{3,j}}\} ,{\boldsymbol{z}_1},{\boldsymbol{z}_2},{\boldsymbol{z}_3}) \!= \!||\left[ \begin{array}{l}
     \!\{{\boldsymbol{x}_{2,j}}\} \\
     \boldsymbol{z}_2 
\end{array} \right]  -\phi_{{\rm{\uppercase\expandafter{\romannumeral2}}}}({\boldsymbol{z}_1}, {\boldsymbol{z}_3}, \{{\boldsymbol{x}_{3,j}}\})||^2 $, where $\phi_{{\rm{\uppercase\expandafter{\romannumeral2}}}}({\boldsymbol{z}_1}, {\boldsymbol{z}_3}, \{{\boldsymbol{x}_{3,j}}\})\!=\! \mathop {\arg \min }\nolimits_{\{ {\boldsymbol{x}_{2,j}}'\} ,{\boldsymbol{z}_2}'} \{ \sum\nolimits_{j = 1}^N {{f_{2,j}}({\boldsymbol{z}_1},{\boldsymbol{x}_{2,j}}',{\boldsymbol{x}_{3,j}})}\!:\! {\boldsymbol{x}_{2,j}}' \!=\! {\boldsymbol{z}_2}', \forall j,  {\boldsymbol{a}_{1,l}^{\rm{\uppercase\expandafter{\romannumeral1}}}}^{\top}\!{\boldsymbol{z}_1} + {\boldsymbol{a}_{2,l}^{\rm{\uppercase\expandafter{\romannumeral1}}}}^{\top}\!{\boldsymbol{z}_2}' + {\boldsymbol{a}_{3,l}^{\rm{\uppercase\expandafter{\romannumeral1}}}}^{\top}\!{\boldsymbol{z}_3} + \sum\nolimits_{j = 1}^N {\boldsymbol{b}{{_{j,l}^{\rm{\uppercase\expandafter{\romannumeral1}}}}}}^{\top}\!{\boldsymbol{x}_{3,j}}  \!\le\! c_l^{\rm{\uppercase\expandafter{\romannumeral1}}},\forall l \}$. 
In Eq. (\ref{eq:4}), the lower-level optimization problem can be viewed as the constraint to the upper-level optimization problem \cite{sinha2017review,gould2016differentiating}, i.e., $h_{{\rm{\uppercase\expandafter{\romannumeral2}}}}(\{{\boldsymbol{x}_{2,j}}\} ,\{ {\boldsymbol{x}_{3,j}}\} ,{\boldsymbol{z}_1},{\boldsymbol{z}_2},{\boldsymbol{z}_3}) =0$.  Likewise, following \cite{jiao2022asynchronous},  the results after $K$ communication rounds between the master and workers are utilized as the estimate of $\phi_{{\rm{\uppercase\expandafter{\romannumeral2}}}}({\boldsymbol{z}_1}, {\boldsymbol{z}_3}, \{{\boldsymbol{x}_{3,j}}\})$. Specifically, for the lower-level optimization problem in Eq. (\ref{eq:4}), the augmented Lagrangian function is given,
\begin{equation}
\begin{array}{l}
  \!   L_{p,2}({\boldsymbol{z}_1},{\boldsymbol{z}_2}', \! \{{\boldsymbol{x}_{2,j}}'\},\! \{ s_l\},\! \{\gamma_l\},\! \{\boldsymbol{\varphi}_{2,j}\}, {\boldsymbol{z}_3},\! \{{\boldsymbol{x}_{3,j}}\}) \vspace{0.5mm} \\
  
\!  =  \sum\nolimits_{j=1}^N  ({{f_{2,j}}({\boldsymbol{z}_1},{\boldsymbol{x}_{2,j}}',{\boldsymbol{x}_{3,j}})} \!+\! {\boldsymbol{\varphi}_{2,j}^{\top}}({\boldsymbol{x}_{2,j}}'\! -\! {\boldsymbol{z}_2}')  \\
\! + \frac{\kappa_2}{2}||{\boldsymbol{x}_{2,j}}'\!-\! {\boldsymbol{z}_2}'||^2 )  
  
+\sum\nolimits_{l = 1}^{|P_{\rm{\uppercase\expandafter{\romannumeral1}}}^{t + 1}|} \gamma_l ( \hat{h}_{\rm{\uppercase\expandafter{\romannumeral1}},l}(\{{\boldsymbol{x}_{3,j}}\},\boldsymbol{z}_1, {\boldsymbol{z}_2}',\boldsymbol{z}_3 )
\\
- c_l^{\rm{\uppercase\expandafter{\romannumeral1}}} \!+\!s_l ) 
   \! + \! \sum\nolimits_{l = 1}^{|P_{\rm{\uppercase\expandafter{\romannumeral1}}}^{t + 1}|} \! \frac{\rho_2}{2} || \hat{h}_{{\rm{\uppercase\expandafter{\romannumeral1}}},l}(\{{\boldsymbol{x}_{3,j}}\},\boldsymbol{z}_1, {\boldsymbol{z}_2}',\boldsymbol{z}_3 ) \! - \! c_l^{\rm{\uppercase\expandafter{\romannumeral1}}} \! + \! s_l||^2,
\end{array}
\end{equation}
where $\gamma_l \!\in\! \mathbb{R}^1$, ${\boldsymbol{\varphi}_{2,j}}\!\in\! \mathbb{R}^{d_2}$ are dual variables, $s_l\!\in\!\mathbb{R}^1_{+},\forall l$ are the slack variables introduced in the inequality constraints, constants $\kappa_2\!>\!0$, $\rho_2\!>\!0$ are penalty parameters. The details of each communication round are presented in Appendix B in the supplementary material.  After $K$ communication rounds, we can obtain the estimate of $\phi_{{\rm{\uppercase\expandafter{\romannumeral2}}}}({\boldsymbol{z}_1}, {\boldsymbol{z}_3}, \{{\boldsymbol{x}_{3,j}}\})$ and the corresponding $h_{{\rm{\uppercase\expandafter{\romannumeral2}}}}$ can be expressed as, 
\begin{equation}
\label{eq:4_23_18}
\begin{array}{l}
 \!    h_{{\rm{\uppercase\expandafter{\romannumeral2}}}}(\{{\boldsymbol{x}_{2,j}}\} ,\{ {\boldsymbol{x}_{3,j}}\} ,{\boldsymbol{z}_1}, {\boldsymbol{z}_2},{\boldsymbol{z}_3}) \\
     
     \!=\! ||\! \left[\! \begin{array}{l}
 \! \{ \boldsymbol{x}_{2,j} - {\boldsymbol{x}_{2,j}^{0}}'+\sum\nolimits_{k = 0}^{K-1} \! \eta_{\boldsymbol{x}}  \nabla_{\boldsymbol{x}_{2,j}}  L_{p,2}^k \} \! \vspace{0.5mm} \\
  
  \boldsymbol{z}_2 - {\boldsymbol{z}_2^{0}}'+\sum\nolimits_{k = 0}^{K-1} \eta_{\boldsymbol{z}} \nabla_{\boldsymbol{z}_2}  L_{p,2}^k
\end{array} \! \right]\!  ||^2 ,
\end{array}
\end{equation}
where $L_{p,2}^k$ is the simplified form of  $ L_{p,2}({\boldsymbol{z}_1},{\boldsymbol{z}_2^k}', \!\{{\boldsymbol{x}_{2,j}^k}'\}, \!\{ s_l^k\},\!\{\gamma_l^k\},\!\{\boldsymbol{\varphi}_{2,j}^k\},\! {\boldsymbol{z}_3}, \!\{{\boldsymbol{x}_{3,j}}\})$. Next, relaxing constraint $h_{{\rm{\uppercase\expandafter{\romannumeral2}}}}(\{{\boldsymbol{x}_{2,j}}\} ,\{ {\boldsymbol{x}_{3,j}}\} ,{\boldsymbol{z}_1},{\boldsymbol{z}_2},{\boldsymbol{z}_3})\!=\!0$ and utilizing \textbf{${\rm{\uppercase\expandafter{\romannumeral2}}}^{\rm{nd}}$ layer} \textbf{polytope} to approximate the feasible region of relaxed constraint $h_{{\rm{\uppercase\expandafter{\romannumeral2}}}}(\{{\boldsymbol{x}_{2,j}}\} ,\{ {\boldsymbol{x}_{3,j}}\} ,{\boldsymbol{z}_1}{\rm{,}}{\boldsymbol{z}_2},{\boldsymbol{z}_3})\!\le\! \varepsilon_{\rm{\uppercase\expandafter{\romannumeral2}}}$. Specifically, the ${\rm{\uppercase\expandafter{\romannumeral2}}}^{\rm{nd}}$ layer polytope can be expressed as $P_{{\rm{\uppercase\expandafter{\romannumeral2}}}}^{t} \!=\! \{\sum\nolimits_{i = 1}^3 {\boldsymbol{a}_{i,l}^{{\rm{\uppercase\expandafter{\romannumeral2}}}}} ^{\top}\!{\boldsymbol{z}_i} + \sum\nolimits_{i = 2}^3 {\sum\nolimits_{j = 1}^N {{\boldsymbol{b}}{{_{i,j,l}^{{\rm{\uppercase\expandafter{\romannumeral2}}}}}}}^{\top}\!{\boldsymbol{x}_{i,j}} }  \!\le\! c_l^{{\rm{\uppercase\expandafter{\romannumeral2}}}},l \!=\! 1,\! \cdots \!,|P_{{\rm{\uppercase\expandafter{\romannumeral2}}}}^{t}|\}$ in $(t+1)^{\rm{th}}$ iteration, where $|P_{{\rm{\uppercase\expandafter{\romannumeral2}}}}^{t}|$ represents the number of cutting planes in $P_{{\rm{\uppercase\expandafter{\romannumeral2}}}}^{t}$, and ${\boldsymbol{a}_{i,l}^{{\rm{\uppercase\expandafter{\romannumeral2}}}}},
{\boldsymbol{b}{{_{i,j,l}^{{\rm{\uppercase\expandafter{\romannumeral2}}}}}}},
c_l^{{\rm{\uppercase\expandafter{\romannumeral2}}}}$ are parameters in $l^{\rm{th}}$ cutting plane (${\rm{\uppercase\expandafter{\romannumeral2}}}^{\rm{nd}}$ layer $\mu$-cut). Thus, the resulting \textbf{hyper-polyhedral} \textbf{approximation} \textbf{problem} is,
\begin{equation}
\label{eq:outer_poly}
\begin{array}{l}
\min \sum\nolimits_{j = 1}^N {{f_{1,j}}({\boldsymbol{x}_{1,j}},{\boldsymbol{x}_{2,j}},{\boldsymbol{x}_{3,j}})} \;{\rm{s}}.{\rm{t}}. \vspace{0.4mm}\\

\; {\boldsymbol{x}_{1,j}} = {\boldsymbol{z}_1},j = 1, \cdots ,N \vspace{0.4mm}\\

\; \sum\nolimits_{i = 1}^3 \! {\boldsymbol{a}_{i,l}^{{\rm{\uppercase\expandafter{\romannumeral2}}}}} ^{\top}\!{\boldsymbol{z}_i} \!+ \!\sum\nolimits_{i = 2}^3 \!{\sum\nolimits_{j = 1}^N \!{\boldsymbol{b}{{_{i,j,l}^{{\rm{\uppercase\expandafter{\romannumeral2}}}}}}}^{\top}\!{\boldsymbol{x}_{i,j}} }  \!\le\! c_l^{{\rm{\uppercase\expandafter{\romannumeral2}}}},l \!=\! 1,\! \cdots \!,|P_{{\rm{\uppercase\expandafter{\romannumeral2}}}}^{t}| \vspace{1mm}\\

{\mathop{\rm var}}.\qquad  \{{\boldsymbol{x}_{1,j}}\} ,{\rm{\{ }}{\boldsymbol{x}_{2,j}}\} ,{\rm{\{ }}{\boldsymbol{x}_{3,j}}\} ,{\boldsymbol{z}_1}{\rm{,}}{\boldsymbol{z}_2},{\boldsymbol{z}_3}.
\end{array}
\end{equation}

It is worth mentioning that solving the TLO problem is theoretically NP-hard (even solving the inner bilevel problem in TLO is NP-hard \cite{ben1990computational}). Thus, it’s unlikely to design a polynomial-time algorithm for the distributed TLO problem unless P = NP \cite{arora2009computational}.  In this work, the hyper-polyhedral
approximation problem in Eq. (\ref{eq:outer_poly}) is a convex relaxation problem of the distributed TLO problem in Eq. (\ref{eq:1}), and the relaxation will be continuously tightened as $\mu$-cuts are added. Detailed discussions are provided in Appendix D.


\subsection{Asynchronous Federated Algorithm}
The synchronous and asynchronous federated algorithms have different application scenarios \cite{su2022gba}. The synchronous algorithm is preferred when the delay of each worker is not much different, and the asynchronous algorithm suits better when there are stragglers in the distributed system. In this work, an asynchronous algorithm is proposed to solve the trilevel optimization problem. Specifically, in the proposed asynchronous algorithm, we set the master
updates its variables once it receives updates from $S (1\le S \le N)$ workers, i.e., active workers, at every iteration, and every
worker has to communicate with the master at least once every $\tau$ iterations to alleviate the staleness issues \cite{zhang2014asynchronous}. It is worth mentioning that $S$ can be flexibly adjusted based on whether there are stragglers, the proposed algorithm becomes synchronous when we set $S=N$, thus the proposed asynchronous algorithm is effective and flexible. First, the Lagrangian function of Eq. (\ref{eq:outer_poly}) can be expressed as,
\begin{equation}
\begin{array}{l}
L_p(\{{\boldsymbol{x}_{1,j}}\} ,\!\{{\boldsymbol{x}_{2,j}}\} ,\!\{{\boldsymbol{x}_{3,j}}\} ,{\boldsymbol{z}_1},{\boldsymbol{z}_2},{\boldsymbol{z}_3},\! \{\lambda_l\}, \!\{\boldsymbol{\theta}_j\} ) \\

= \sum\nolimits_{j = 1}^N \! {{f_{1,j}}({\boldsymbol{x}_{1,j}},{\boldsymbol{x}_{2,j}},{\boldsymbol{x}_{3,j}})} \! +\! \sum\nolimits_{j = 1}^N \!{{\boldsymbol{\theta}_j}}^{\top} ({\boldsymbol{x}_{1,j}}\! -\! {\boldsymbol{z}_1})\\

 + \sum\nolimits_{l = 1}^{|P_{{\rm{\uppercase\expandafter{\romannumeral2}}}}^{t}|} {{\lambda _l}} ({\sum\nolimits_{i = 1}^3 {\boldsymbol{a}_{i,l}^{{\rm{\uppercase\expandafter{\romannumeral2}}}}} ^{\top}}{\boldsymbol{z}_i} \!+\! \sum\nolimits_{i = 2}^3 \! {\sum\nolimits_{j = 1}^N {\boldsymbol{b}{{_{i,j,l}^{{\rm{\uppercase\expandafter{\romannumeral2}}}}}^{\top}}{\boldsymbol{x}_{i,j}}} } \! -\! c_l^{{\rm{\uppercase\expandafter{\romannumeral2}}}}),
\end{array}
\end{equation}
where $\lambda _l\!\in\!\mathbb{R}^1_{+}$, $\boldsymbol{\theta}_j\!\in\!\mathbb{R}^{d_1}$ are dual variables. Following \cite{xu2020unified,jiao2022asynchronous}, the regularized Lagrangian function is used to update variables as follows,
\begin{equation}
\begin{array}{l}
\widehat{L}_p(\{{\boldsymbol{x}_{1,j}}\} ,\!\{{\boldsymbol{x}_{2,j}}\} ,\!\{{\boldsymbol{x}_{3,j}}\} ,{\boldsymbol{z}_1},{\boldsymbol{z}_2},{\boldsymbol{z}_3},\! \{\lambda_l\}, \!\{\boldsymbol{\theta}_j\} ) \\ = {L_p} - \sum\nolimits_{l = 1}^{|P_{{\rm{\uppercase\expandafter{\romannumeral2}}}}^{t}|} \!\frac{c_1^t}{2}||\lambda_l||^2 - \sum\nolimits_{j = 1}^{N}\! \frac{c_2^t}{2}||\boldsymbol{\theta}_j||^2,
\end{array}
\end{equation}
where $L_p\!=\!L_p(\{{\boldsymbol{x}_{1,j}}\} ,\!\{{\boldsymbol{x}_{2,j}}\} ,\!\{{\boldsymbol{x}_{3,j}}\} ,\!{\boldsymbol{z}_1},\!{\boldsymbol{z}_2},\!{\boldsymbol{z}_3},\! \{\lambda_l\}, \!\{\boldsymbol{\theta}_j\} )$, and $c_1^t$, $c_2^t$ are the regularization terms in $(t+1)^{\rm{th}}$ iteration. We set that ${c_1^t} =1/{{{\eta _{{\lambda}}}}{(t+1)^{\frac{1}{4}}}} \ge  \underline{c}_1$, ${c_2^t} = 1/{{{\eta _{{\boldsymbol{\theta }}}}}{(t+1)^{\frac{1}{4}}}}  \ge  \underline{c}_2$ are two nonnegative non-increasing sequences, where $\eta _{{\lambda}}$, $\eta _{{\boldsymbol{\theta }}}$, $\underline{c}_1$, $\underline{c}_2$ are constants, and $\underline{c}_1$, $\underline{c}_2$ meet that $0\!<\!\underline{c}_1\!<\!1/\eta _{{\lambda}}({({{4M\alpha_4}} / {{{{\eta _{\lambda}}}^2}} \!+\! {{4N\alpha_5}}/{{{{\eta _{\boldsymbol{\theta }}}}^2}})}1/\epsilon)^{\frac{1}{2}}$ and $0\!<\!\underline{c}_2\!<\!1/\eta _{{\boldsymbol{\theta }}} ({({{4M\alpha_4}} / {{{{\eta _{\lambda}}}^2}} \!+\! {{4N\alpha_5}}/{{{{\eta _{\boldsymbol{\theta }}}}^2}})}1/\epsilon)^{\frac{1}{2}}$ ($\epsilon$ refers to the tolerance error, and $\alpha_4$, $\alpha_5$ are constants, which will be introduced below). In $(t+1)^{\rm{th}}$ master iteration, $Q^{t+1} $ is utilized to denote the index set of active workers, and the proposed asynchronous algorithm proceeds as follows,

\noindent (1) \emph{Active} \emph{workers} update the local variables as follows,
\begin{equation}
\label{eq:update_x1_asyn}
{\boldsymbol{x}_{i,j}^{t+1}} = \left\{ \begin{array}{l}
{\boldsymbol{x}_{i,j}^{t}} - \eta_{\boldsymbol{x}_i} \nabla_{\!\boldsymbol{x}_{i,j}} \widehat{L}_p^{\hat{t}_j}, j\in Q^{t+1}\\
{\boldsymbol{x}_{i,j}^{t}}, j\notin Q^{t+1}
\end{array} \right., \forall i,
\end{equation}
where $\eta_{\boldsymbol{x}_i} (\forall i=1,2,3)$ denote the step-sizes, $\widehat{L}_p^{\hat{t}_j} = \widehat{L}_p(\{{\boldsymbol{x}_{i,j}^{\hat{t}_j}}\} ,\!\{{\boldsymbol{z}_i^{\hat{t}_j}}\}, \!\{\lambda_l^{\hat{t}_j}\}, \!\{\boldsymbol{\theta}_j^{\hat{t}_j}\} )$ and ${\hat{t}_j}$ denotes the last iteration that worker $j$ is active. Then, active workers (i.e., worker $j, j\in Q^{t+1}$) transmit the updated local variables, i.e., $\boldsymbol{x}_{i,j}^{t+1}, \forall i$ to the master.

\noindent (2) After receiving the updates from workers, the \emph{master} updates the variables as follows,
\begin{equation}
\label{eq:update_z1_asyn}
\!{\boldsymbol{z}_1^{t+1}} \!=\! {\boldsymbol{z}_1^{t}}\!-\!\eta_{\boldsymbol{z}_1}\! \nabla_{\boldsymbol{z}_1}\widehat{L}_p(\{{\boldsymbol{x}_{i,j}^{t+1}}\} ,\!\{{\boldsymbol{z}_i^t} \},\! \{\lambda_l^t\},\! \{\boldsymbol{\theta}_j^t\} ),
\end{equation}
\begin{equation}
\label{eq:update_z2_asyn}
{\boldsymbol{z}_2^{t+1}} \!=\! {\boldsymbol{z}_2^{t}} \!-\!\eta_{\boldsymbol{z}_2}\! \nabla_{\boldsymbol{z}_2}\widehat{L}_p(\{{\boldsymbol{x}_{i,j}^{t+1}}\} ,\!{\boldsymbol{z}_1^{t+1}}\!,{\boldsymbol{z}_2^t},{\boldsymbol{z}_3^t},\! \{\lambda_l^t\},\! \{\boldsymbol{\theta}_j^t\} ),
\end{equation}
\begin{equation}
\label{eq:update_z3_asyn}
{\boldsymbol{z}_3^{t+1}} \!=\! {\boldsymbol{z}_3^{t}}\!-\!\eta_{\boldsymbol{z}_3} \! \nabla_{\boldsymbol{z}_3}\widehat{L}_p(\{{\boldsymbol{x}_{i,j}^{t+1}}\} ,\!{\boldsymbol{z}_1^{t+1}}\!,{\boldsymbol{z}_2^{t+1}}\!,{\boldsymbol{z}_3^t},\! \{\lambda_l^t\},\! \{\boldsymbol{\theta}_j^t\} ),
\end{equation}
\begin{equation}
\label{eq:update_lambda_asyn}
\!{\lambda_l^{t+1}} \!=\! \mathcal{P}_{\Lambda} ( {\lambda_l^{t}}\!+\!\eta_{\lambda}\! \nabla_{\lambda_l}\widehat{L}_p(\{{\boldsymbol{x}_{i,j}^{t+1}}\} ,\!\{{\boldsymbol{z}_i^{t+1}}\},\! \{\lambda_l^t\},\! \{\boldsymbol{\theta}_j^t\} ) ),
\end{equation}
\begin{equation}
\label{eq:update_theta_asyn}
\!{\boldsymbol{\theta}_j^{t+1}} \!=\! \mathcal{P}_{\Theta}({\boldsymbol{\theta}_j^{t}}\!+\!\eta_{\boldsymbol{\theta}}\! \nabla_{\boldsymbol{\theta}_j}\widehat{L}_p(\{{\boldsymbol{x}_{i,j}^{t+1}}\} ,\!\{{\boldsymbol{z}_i^{t+1}}\}, \!\{\lambda_l^{t+1}\}, \!\{\boldsymbol{\theta}_j^t\} )),
\end{equation}
where $\eta_{\boldsymbol{z}_1}, \eta_{\boldsymbol{z}_2}, \eta_{\boldsymbol{z}_3}, \eta_{\lambda}, \eta_{\boldsymbol{\theta}}$ denote the step-sizes, $\mathcal{P}_{\Lambda}$ and $\mathcal{P}_{\Theta}$ represent the projection onto sets ${\Lambda}=\{ \lambda_l|\; 0 \! \le \! \lambda_l \! \le \! \sqrt{\alpha_4}  \}$ and ${\Theta}=\{ \boldsymbol{\theta}_j|\; ||\boldsymbol{\theta}_j||_{\infty}\! \le \! \sqrt{\alpha_5}/d_1 \}$, where $\alpha_4 \!>\!0$ and $\alpha_5 \!>\!0$ are constants. Then, master broadcasts the updated variables, i.e., ${\boldsymbol{z}_1^{t+1}}, {\boldsymbol{z}_2^{t+1}}, {\boldsymbol{z}_3^{t+1}}, \{\lambda_l^{t+1}\}, \boldsymbol{\theta}_j$ to the active worker $j$. Details are summarized in Algorithm \ref{algorithm:asyn}.

\begin{algorithm}[t]
   \caption{Asynchronous Federated Trilevel Learning}
\begin{algorithmic}
   \STATE {\bfseries Initialization:}  master iteration $t\! =\! 0$, variables $\{{{\boldsymbol{x}}_{1,j}^0}\},$ $ \{{{\boldsymbol{x}}_{2,j}^0}\}, $ $ \{{{\boldsymbol{x}}_{3,j}^0}\}, \boldsymbol{z}_1^0, \boldsymbol{z}_2^0, \boldsymbol{z}_3^0, \{\lambda_l^0\}, \{\boldsymbol{\theta}_j^0\}$.
   
   \REPEAT
   
   \FOR{\emph{active worker}}
   \STATE updates variables $\boldsymbol{x}_{1,j}^{t+1}$, $\boldsymbol{x}_{2,j}^{t+1}$ and $\boldsymbol{x}_{3,j}^{t+1}$ by Eq. (\ref{eq:update_x1_asyn});
   \ENDFOR
   
   \STATE active workers send updated local variables to master;
   
   \FOR{\emph{master}}
   \STATE updates variables $\boldsymbol{z}_1^{t+1}, \boldsymbol{z}_2^{t+1}, \boldsymbol{z}_3^{t+1}, \{\lambda_l^{t+1}\}, \{\boldsymbol{\theta}_j^{t+1}\}$  by Eq. (\ref{eq:update_z1_asyn}), (\ref{eq:update_z2_asyn}), (\ref{eq:update_z3_asyn}), (\ref{eq:update_lambda_asyn}) and (\ref{eq:update_theta_asyn});
   \ENDFOR
   \STATE master broadcasts updated variables to active workers;

   \IF{$(t+1)$ mod $T_{\rm{pre}}$ $==0$ and $t<T_1$}
   \STATE   new ${\rm{\uppercase\expandafter{\romannumeral1}}}^{\rm{st}}$ layer $\mu$-cut ${cp_{\rm{\uppercase\expandafter{\romannumeral1}}}}$ is generated by Eq. (\ref{eq:new_inner_cp}) and added into ${\rm{\uppercase\expandafter{\romannumeral1}}}^{\rm{st}}$ layer polytope;
   \STATE   new ${\rm{\uppercase\expandafter{\romannumeral2}}}^{\rm{nd}}$ layer $\mu$-cut ${cp_{\rm{\uppercase\expandafter{\romannumeral2}}}}$ is generated by Eq. (\ref{eq:new_outer_cp}) and added into ${\rm{\uppercase\expandafter{\romannumeral2}}}^{\rm{nd}}$ layer polytope;

   \STATE removing inactive ${\rm{\uppercase\expandafter{\romannumeral1}}}^{\rm{st}}$,  ${\rm{\uppercase\expandafter{\romannumeral2}}}^{\rm{nd}}$ layer $\mu$-cuts by Eq. (\ref{eq:inactive_inner});
   \ENDIF

   \STATE $t =t+1$;
   \UNTIL{termination.}

\end{algorithmic}
\label{algorithm:asyn}
\end{algorithm}

\subsection{Refining Hyper-polyhedral Approximation}

In this section, a novel $\mu$-cut is proposed, which can be utilized for non-convex ($\mu$-weakly convex) optimization problem and thus is more general than the traditional cutting plane designed for convex optimization  \cite{jiao2022asynchronous,franc2011cutting}. We demonstrate that the proposed $\mu$-cuts are valid, i.e., the original feasible region is a subset of the polytope that forms of $\mu$-cuts in Proposition \ref{prop:1} and \ref{prop:2}.  Every $T_{\rm{pre}}$ iteration, the $\mu$-cuts will be updated to refine the hyper-polyhedral approximation when $t<T_1$, which can be divided into three steps: 1) generating new ${\rm{\uppercase\expandafter{\romannumeral1}}}^{\rm{st}}$ layer $\mu$-cut, 2) generating new ${\rm{\uppercase\expandafter{\romannumeral2}}}^{\rm{nd}}$ layer $\mu$-cut, 3) removing inactive $\mu$-cuts.

\subsubsection{Generating new ${\rm{\uppercase\expandafter{\romannumeral1}}}^{\rm{st}}$ layer $\mu$-cut:}

Following \cite{qian2019robust}, we assume the variables are bounded, i.e., $||\boldsymbol{x}_{i,j}||^2\!\le\!\alpha_i$, $||\boldsymbol{z}_i||^2\!\le\!\alpha_i,i\!=\!1,2,3$, and $h_{\rm{\uppercase\expandafter{\romannumeral1}}}$ is $\mu$-weakly convex. It is demonstrated in Appendix E that  $h_{\rm{\uppercase\expandafter{\romannumeral1}}}$ is $\mu$-weakly convex in lots of cases. Following \cite{xie2019asynchronous,davis2019stochastic}, the definition and first-order condition of $\mu$-weakly convex function are given as follows.

\begin{definition}
\textbf{($\mu$-weakly convex)} A differentiable function $f(\boldsymbol{x})$ is $\mu$-weakly convex if function $g(\boldsymbol{x})=f(\boldsymbol{x})+\frac{\mu}{2}||\boldsymbol{x}||^2$ is convex. 
\end{definition}

\begin{definition}
\label{def:2}
\textbf{(First-order condition)} For any $\boldsymbol{x}$, $\boldsymbol{x}'$, a differentiable function $f(\boldsymbol{x})$ is $\mu$-weakly convex if and only if the following inequality holds.
\begin{equation}
f(\boldsymbol{x})\ge f(\boldsymbol{x}') + \nabla f(\boldsymbol{x}')^{\top}(\boldsymbol{x}-\boldsymbol{x}') - \frac{\mu}{2} ||\boldsymbol{x}-\boldsymbol{x}'||^2.
\end{equation}
\end{definition}

Combining the first-order condition of $\mu$-weakly convex function with Cauchy-Schwarz inequality, a kind of new cutting plane, i.e., ${\mu}$-\textbf{cut}, can be generated. Specifically, the new ${\rm{\uppercase\expandafter{\romannumeral1}}}^{\rm{st}}$ layer $\mu$-cut ${cp_{\rm{\uppercase\expandafter{\romannumeral1}}}}$ for ${\rm{\uppercase\expandafter{\romannumeral1}}}^{\rm{st}}$ layer polytope can be expressed as:
\begin{equation}
\label{eq:new_inner_cp}
\begin{array}{l}
 {\left[ \! \begin{array}{l}
\{\frac{{\partial h_{\rm{\uppercase\expandafter{\romannumeral1}}}(\{{\boldsymbol{x}_{3,j}^{t+1}}\},\boldsymbol{z}_1^{t+1}, {\boldsymbol{z}_2^{t+1}}',\boldsymbol{z}_3^{t+1} )}}{{\partial {\boldsymbol{x}_{3,j}}}} \}\\
\, \, \frac{{\partial h_{\rm{\uppercase\expandafter{\romannumeral1}}}(\{{\boldsymbol{x}_{3,j}^{t+1}}\},\boldsymbol{z}_1^{t+1}, {\boldsymbol{z}_2^{t+1}}',\boldsymbol{z}_3^{t+1} )}}{{\partial \boldsymbol{z}_1}} \\
\, \,  \frac{{\partial h_{\rm{\uppercase\expandafter{\romannumeral1}}}(\{{\boldsymbol{x}_{3,j}^{t+1}}\},\boldsymbol{z}_1^{t+1}, {\boldsymbol{z}_2^{t+1}}',\boldsymbol{z}_3^{t+1} )}}{{\partial {\boldsymbol{z}_2}'}} \\
\, \,  \frac{{\partial h_{\rm{\uppercase\expandafter{\romannumeral1}}}(\{{\boldsymbol{x}_{3,j}^{t+1}}\},\boldsymbol{z}_1^{t+1}, {\boldsymbol{z}_2^{t+1}}',\boldsymbol{z}_3^{t+1} )}}{{\partial \boldsymbol{z}_3}} 
\end{array} \! \right]^{\top}}\! \left[\! \begin{array}{l}
\{ \boldsymbol{x}_{3,j} - \boldsymbol{x}_{3,j}^{t+1} \}\\
\; \,  \boldsymbol{z}_1 - \boldsymbol{z}_1^{t+1}  \\
\; \,  {\boldsymbol{z}_2}' - {\boldsymbol{z}_2^{t+1}}' \\
\; \,  \boldsymbol{z}_3 - \boldsymbol{z}_3^{t+1} 
\end{array} \! \right] \\

\!+h_{\rm{\uppercase\expandafter{\romannumeral1}}}(\{{\boldsymbol{x}_{3,j}^{t+1}}\},\boldsymbol{z}_1^{t+1}, {\boldsymbol{z}_2^{t+1}}',\boldsymbol{z}_3^{t+1} ) 
\! \le \! \varepsilon_{\rm{\uppercase\expandafter{\romannumeral1}}} \! +\! \mu ((N\!+\!1)\alpha_1 \!+ \! \alpha_2
\\
\!+\alpha_3  \!+\! \sum\nolimits_{j = 1}^N \!||\boldsymbol{x}_{3,j}^{t+1}||^2 \!+\! ||\boldsymbol{z}_1^{t+1}||^2 \!+\! ||{\boldsymbol{z}_2^{t+1}}'||^2 \!+\! ||\boldsymbol{z}_3^{t+1}||^2).
\end{array}
\end{equation}

\begin{prop}
\label{prop:1}
The feasible region of constraint $h_{\rm{\uppercase\expandafter{\romannumeral1}}}(\{{\boldsymbol{x}_{3,j}}\},\!\boldsymbol{z}_1, {\boldsymbol{z}_2}',\boldsymbol{z}_3 )\! \le \! \varepsilon_{\rm{\uppercase\expandafter{\romannumeral1}}}$  is a subset of the ${\rm{\uppercase\expandafter{\romannumeral1}}}^{\rm{st}}$ layer polytope  $P_{\rm{\uppercase\expandafter{\romannumeral1}}}^{t}\!=\! \{{\boldsymbol{a}_{1,l}^{\rm{\uppercase\expandafter{\romannumeral1}}}}^{\top}\!{\boldsymbol{z}_1} + {\boldsymbol{a}_{2,l}^{\rm{\uppercase\expandafter{\romannumeral1}}}}^{\top}\!{\boldsymbol{z}_2}' + {\boldsymbol{a}_{3,l}^{\rm{\uppercase\expandafter{\romannumeral1}}}}^{\top}\!{\boldsymbol{z}_3} + \sum\nolimits_{j = 1}^N {\boldsymbol{b}{{_{j,l}^{\rm{\uppercase\expandafter{\romannumeral1}}}}}}^{\top}\!{\boldsymbol{x}_{3,j}}  \!\le\! c_l^{\rm{\uppercase\expandafter{\romannumeral1}}},l \!=\! 1,\! \cdots\! ,|P_{\rm{\uppercase\expandafter{\romannumeral1}}}^{t}| \}$. In addition, $P_{\rm{\uppercase\expandafter{\romannumeral1}}}^{t}$ converges monotonically with the number of $\mu$-cuts. The proof is given in Appendix C.
\end{prop}

Note that if $h_{\rm{\uppercase\expandafter{\romannumeral1}}}$ is convex, i.e., $\mu=0$, the cutting plane will be generated as the same as that in \cite{franc2011cutting,jiao2022asynchronous}, which is designed for convex optimization. Thus, the proposed $\mu$-cut is more general than prior work in the literature. Consequently, the ${\rm{\uppercase\expandafter{\romannumeral1}}}^{\rm{st}}$ layer polytope will be updated as
${P_{\rm{\uppercase\expandafter{\romannumeral1}}}^{t + 1}} = {\rm{Add}}({P_{\rm{\uppercase\expandafter{\romannumeral1}}}^{t}},{cp_{\rm{\uppercase\expandafter{\romannumeral1}}}})$, 
where ${\rm{Add}}({P_{\rm{\uppercase\expandafter{\romannumeral1}}}^{t}},{cp_{\rm{\uppercase\expandafter{\romannumeral1}}}})$ represents adding new $\mu$-cut ${cp_{\rm{\uppercase\expandafter{\romannumeral1}}}}$ into the polytope ${P_{\rm{\uppercase\expandafter{\romannumeral1}}}^{t}}$.

\subsubsection{Generating new ${\rm{\uppercase\expandafter{\romannumeral2}}}^{\rm{nd}}$ layer $\mu$-cut:}

Based on the updated ${\rm{\uppercase\expandafter{\romannumeral1}}}^{\rm{st}}$ layer polytope, the ${\rm{\uppercase\expandafter{\romannumeral2}}}^{\rm{nd}}$ layer polytope will be updated.  The new generated ${\rm{\uppercase\expandafter{\romannumeral2}}}^{\rm{nd}}$ layer $\mu$-cut ${cp_{\rm{\uppercase\expandafter{\romannumeral2}}}}$ can be written as,
\begin{equation}
\label{eq:new_outer_cp}
\begin{array}{l}
\! {\left[ \! \begin{array}{l}
\!\{\frac{{\partial h_{{\rm{\uppercase\expandafter{\romannumeral2}}}}(\{{\boldsymbol{x}_{2,j}^{t+1}}\} ,\{ {\boldsymbol{x}_{3,j}^{t+1}}\} ,{\boldsymbol{z}_1^{t+1}}{\rm{,}}{\boldsymbol{z}_2^{t+1}},{\boldsymbol{z}_3^{t+1}})}}{{\partial {\boldsymbol{x}_{2,j}}}} \}\!\\
\!\{\frac{{\partial h_{{\rm{\uppercase\expandafter{\romannumeral2}}}}(\{{\boldsymbol{x}_{2,j}^{t+1}}\} ,\{ {\boldsymbol{x}_{3,j}^{t+1}}\} ,{\boldsymbol{z}_1^{t+1}}{\rm{,}}{\boldsymbol{z}_2^{t+1}},{\boldsymbol{z}_3^{t+1}})}}{{\partial {\boldsymbol{x}_{3,j}}}} \}\!\\
\, \, \frac{{\partial h_{{\rm{\uppercase\expandafter{\romannumeral2}}}}(\{{\boldsymbol{x}_{2,j}^{t+1}}\} ,\{ {\boldsymbol{x}_{3,j}^{t+1}}\} ,{\boldsymbol{z}_1^{t+1}}{\rm{,}}{\boldsymbol{z}_2^{t+1}},{\boldsymbol{z}_3^{t+1}})}}{{\partial \boldsymbol{z}_1}} \\
\, \,  \frac{{\partial h_{{\rm{\uppercase\expandafter{\romannumeral2}}}}(\{{\boldsymbol{x}_{2,j}^{t+1}}\} ,\{ {\boldsymbol{x}_{3,j}^{t+1}}\} ,{\boldsymbol{z}_1^{t+1}}{\rm{,}}{\boldsymbol{z}_2^{t+1}},{\boldsymbol{z}_3^{t+1}})}}{{\partial \boldsymbol{z}_2}} \\
\, \,  \frac{{\partial h_{{\rm{\uppercase\expandafter{\romannumeral2}}}}(\{{\boldsymbol{x}_{2,j}^{t+1}}\} ,\{ {\boldsymbol{x}_{3,j}^{t+1}}\} ,{\boldsymbol{z}_1^{t+1}}{\rm{,}}{\boldsymbol{z}_2^{t+1}},{\boldsymbol{z}_3^{t+1}})}}{{\partial \boldsymbol{z}_3}} 
\end{array} \! \right]^{\top}}\! \left[\! \begin{array}{l}
\! \{ \boldsymbol{x}_{2,j} \!-\! \boldsymbol{x}_{2,j}^{t+1} \}\!\\
\!\{ \boldsymbol{x}_{3,j} \!-\! \boldsymbol{x}_{3,j}^{t+1} \}\!\\
\; \,  \boldsymbol{z}_1 - \boldsymbol{z}_1^{t+1}  \\
\; \,  \boldsymbol{z}_2 - {\boldsymbol{z}_2^{t+1}} \\
\; \,  \boldsymbol{z}_3 - \boldsymbol{z}_3^{t+1} 
\end{array} \! \right] \\

\!+h_{{\rm{\uppercase\expandafter{\romannumeral2}}}}(\{{\boldsymbol{x}_{2,j}^{t+1}}\} ,\!\{ {\boldsymbol{x}_{3,j}^{t+1}}\} ,\!{\boldsymbol{z}_1^{t+1}}{\rm{,}}{\boldsymbol{z}_2^{t+1}},\!{\boldsymbol{z}_3^{t+1}}) 
\le \varepsilon_{\rm{\uppercase\expandafter{\romannumeral2}}}  + \mu (\alpha_1 \\
\!+ (N\!+\!1)(\alpha_2\!+\!\alpha_3) \!+\! \sum\nolimits_{i = 2}^3 \! \sum\nolimits_{j = 1}^N \!||\boldsymbol{x}_{i,j}^{t+1}||^2 \!+\! \sum\nolimits_{i = 1}^3 \!||\boldsymbol{z}_i^{t+1}||^2 ).
\end{array}
\end{equation}

Consequently, the ${\rm{\uppercase\expandafter{\romannumeral2}}}^{\rm{nd}}$ layer polytope will be updated as 
$ {P_{{\rm{\uppercase\expandafter{\romannumeral2}}}}^{t + 1}} = {\rm{Add}}({P_{{\rm{\uppercase\expandafter{\romannumeral2}}}}^{t}},{cp_{\rm{\uppercase\expandafter{\romannumeral2}}}}).$

\begin{prop}
\label{prop:2}
The feasible region of constraint $h_{{\rm{\uppercase\expandafter{\romannumeral2}}}}(\{{\boldsymbol{x}_{2,j}}\} ,\!\{ {\boldsymbol{x}_{3,j}}\} ,\!\{{\boldsymbol{z}_i}\})\! \le \! \varepsilon_{\rm{\uppercase\expandafter{\romannumeral2}}}$  is a subset of the ${\rm{\uppercase\expandafter{\romannumeral2}}}^{\rm{nd}}$ layer polytope  $P_{\rm{\uppercase\expandafter{\romannumeral2}}}^{t}\!=\!\{\sum\nolimits_{i = 1}^3 \!{\boldsymbol{a}_{i,l}^{{\rm{\uppercase\expandafter{\romannumeral2}}}}} ^{\top}\!{\boldsymbol{z}_i} \!+\! \sum\nolimits_{i = 2}^3 \!{\sum\nolimits_{j = 1}^N \!{\boldsymbol{b}{{_{i,j,l}^{{\rm{\uppercase\expandafter{\romannumeral2}}}}}}}^{\top}\!{\boldsymbol{x}_{i,j}} }  \!\le\! c_l^{{\rm{\uppercase\expandafter{\romannumeral2}}}},l \!=\! 1,\! \cdots \!,|P_{{\rm{\uppercase\expandafter{\romannumeral2}}}}^{t}| \}$, and  $P_{\rm{\uppercase\expandafter{\romannumeral2}}}^{t}$ converges monotonically with the number of $\mu$-cuts. The proof is given in Appendix C.
\end{prop}

\subsubsection{Removing inactive $\mu$-cuts:}

Removing the inactive cutting planes can enhance the efficiency of the proposed algorithm \cite{yang2014distributed,jiao2022distributed}. The inactive ${\rm{\uppercase\expandafter{\romannumeral1}}}^{\rm{st}}$ and ${\rm{\uppercase\expandafter{\romannumeral2}}}^{\rm{nd}}$ layer $\mu$-cuts will be removed, thus the corresponding ${\rm{\uppercase\expandafter{\romannumeral1}}}^{\rm{st}}$ and ${\rm{\uppercase\expandafter{\romannumeral2}}}^{\rm{nd}}$ layer polytopes ${P_{\rm{\uppercase\expandafter{\romannumeral1}}}^{t + 1}}$ and ${P_{\rm{\uppercase\expandafter{\romannumeral2}}}^{t + 1}}$ will be updated as follows.
\begin{equation}
\begin{array}{l}
\label{eq:inactive_inner}
{P_{\rm{\uppercase\expandafter{\romannumeral1}}}^{t + 1}} = \left\{ \begin{array}{l}
{\rm{Drop}}({P_{\rm{\uppercase\expandafter{\romannumeral1}}}^{t + 1}},c{p_l^{\rm{\uppercase\expandafter{\romannumeral1}}}}),{\rm{if}} \;\gamma_l^K=0\\
{P_{\rm{\uppercase\expandafter{\romannumeral1}}}^{t + 1}},{\rm{otherwise}}
\end{array} \right. , \vspace{1mm}\\

{P_{{\rm{\uppercase\expandafter{\romannumeral2}}}}^{t + 1}} = \left\{ \begin{array}{l}
{\rm{Drop}}({P_{{\rm{\uppercase\expandafter{\romannumeral2}}}}^{t + 1}},c{p_l^{\rm{\uppercase\expandafter{\romannumeral2}}}}),{\rm{if}} \;\lambda_l^{t+1}=0\\
{P_{{\rm{\uppercase\expandafter{\romannumeral2}}}}^{t + 1}},{\rm{otherwise}}
\end{array} \right. ,
\end{array}
\end{equation}
where ${\rm{Drop}}(P,c{p_l})$ represents that the $l^{\rm{th}}$ cutting plane $c{p_l}$ is removed from polytope $P$.

\section{Discussion}
\begin{definition}
\label{definition:1}
{\textbf{(Stationarity gap)}} Following \cite{xu2020unified,lu2020hybrid,jiao2022distributed}, the \textit{stationarity} \textit{gap} of our problem at $t^{{th}}$ iteration is defined as:
\begin{equation}
\label{eq:definition1}
\nabla G^t = \left[ \begin{array}{l}
\{ {\nabla _{\boldsymbol{x}_{i,j}}}L_p(\{{\boldsymbol{x}_{i,j}^t}\} ,\!\{{\boldsymbol{z}_i^t} \},\! \{\lambda_l^t\}, \!\{\boldsymbol{\theta}_j^t\} )\}\\

\{ {\nabla _{\boldsymbol{z}_i}}L_p(\{{\boldsymbol{x}_{i,j}^t}\} ,\!\{{\boldsymbol{z}_i^t} \},\! \{\lambda_l^t\}, \!\{\boldsymbol{\theta}_j^t\} ) \} \\

\{  {\nabla G_{{\lambda_l}}}(\{{\boldsymbol{x}_{i,j}^t}\} ,\!\{{\boldsymbol{z}_i^t} \},\! \{\lambda_l^t\}, \!\{\boldsymbol{\theta}_j^t\} )   \} \\
\{{\nabla G_{{\boldsymbol{\theta}_j}}}(\{{\boldsymbol{x}_{i,j}^t}\} ,\!\{{\boldsymbol{z}_i^t} \},\! \{\lambda_l^t\}, \!\{\boldsymbol{\theta}_j^t\} )\} \\
\end{array} \right],
\end{equation}
\end{definition}
where 
\begin{equation}
\begin{array}{l}
{\nabla G_{{\lambda_l}}}(\{{\boldsymbol{x}_{i,j}^t}\},\!\{{\boldsymbol{z}_i^t} \},\! \{\lambda_l^t\}, \!\{\boldsymbol{\theta}_j^t\} )
   
   \\ \!= \! \frac{1}{\eta_{\lambda}}\left(\lambda_l^t \!-\!\mathcal{P}_{\Lambda} (\lambda_l^t \!+\! \eta_{\lambda}{\nabla _{\lambda_l}}L_p(\{{\boldsymbol{x}_{i,j}^t}\} ,\!\{{\boldsymbol{z}_i^t} \},\! \{\lambda_l^t\}, \!\{\boldsymbol{\theta}_j^t\} ))\right) , \vspace{1mm} \\

{\nabla G_{{\boldsymbol{\theta}_j}}}(\{{\boldsymbol{x}_{i,j}^t}\},\!\{{\boldsymbol{z}_i^t} \},\! \{\lambda_l^t\}, \!\{\boldsymbol{\theta}_j^t\} )
   
   \\ \! = \! \frac{1}{\eta_{\boldsymbol{\theta}}}\left(\boldsymbol{\theta}_j^t \!- \! \mathcal{P}_{\Theta} (\boldsymbol{\theta}_j^t \!+\! \eta_{\boldsymbol{\theta}}{\nabla _{\boldsymbol{\theta}_j}}L_p(\{{\boldsymbol{x}_{i,j}^t}\} ,\!\{{\boldsymbol{z}_i^t} \},\! \{\lambda_l^t\}, \!\{\boldsymbol{\theta}_j^t\} ))\right) .
\end{array}
\end{equation}

\begin{definition}
\label{definition:e stationary point}
{\textbf{($\epsilon$-stationary point)}} If $\,||\nabla G^t||^2 \le \epsilon $,  $(\{{\boldsymbol{x}_{i,j}^t}\} ,\! \{{\boldsymbol{z}_i^t}\},\! \{\lambda_l^t\}, \!\{\boldsymbol{\theta}_j^t\} )$ is an $\epsilon$-stationary point ($\epsilon  \ge 0$) of a differentiable function ${L_p}$.  $T(\epsilon )$ is the first iteration index such that $||\nabla G^t||^2   \le  \epsilon$, \textit{i.e.}, $T(\epsilon )  =  \min \{ t \ |\; ||\nabla G^t||^2   \le  \epsilon \}  $.
\end{definition}

\begin{assumption}\label{assumption:1}
\textbf{{{(Gradient Lipschitz})}} Following \cite{ji2021bilevel}, we assume that
$L_p$ has Lipschitz continuous gradients, i.e., for any $\boldsymbol{\omega}, \boldsymbol{\omega}'$, we assume that there exists $L>0$ satisfying that,
\begin{equation}
    \begin{array}{l}
||{\nabla}{L_p}(\boldsymbol{\omega})  -  {\nabla}{L_p}(\boldsymbol{\omega}')|| \le L||\boldsymbol{\omega}-\boldsymbol{\omega}'||.
\end{array}
\end{equation}
\end{assumption}

\begin{assumption}\label{assumption:2}
{\textbf{(Boundedness)}} Following \cite{qian2019robust,jiao2022asynchronous}, we assume $||\boldsymbol{x}_{i,j}||^2\!\le\!\alpha_i$, $||\boldsymbol{z}_i||^2\!\le\!\alpha_i,i=1,2,3$. And we assume that before obtaining the $\epsilon$-stationary point (i.e., $t\!\le\! T(\epsilon )\!-\!1$), the variables in master satisfy that $\sum\nolimits_i||\boldsymbol{z}_i^{t+1} \!-\! \boldsymbol{z}_i^t|{|^2}\!+\!\sum\nolimits_l||{\lambda _l^{t + 1}} \!-\! {\lambda _l^t}|{|^2} \ge \vartheta $, where $\vartheta >0$ is a relative small constant. The change of the variables in master is upper bounded within $\tau$ iterations:
\begin{equation}
\begin{array}{*{20}{l}}
\!{\!||\boldsymbol{z}_i^t \!-\! \boldsymbol{z}_i^{t - k}|{|^2} \! \le \! \tau{k_1}\vartheta},
{\sum\nolimits_l \! ||{\lambda _l^t} \!-\! {\lambda _l^{t - k}}|{|^2} \! \le \! \tau{k_1}\vartheta}, {1 \! \le \! k \! \le \! \tau },
\end{array}
\end{equation}
where $k_1 >0$ is a constant. Detailed discussions about Assumption \ref{assumption:1} and \ref{assumption:2} are provided in Appendix I.
\end{assumption}

\begin{theorem}
\label{theorem 2}
{\textbf{(Iteration Complexity)}} Suppose Assumption \ref{assumption:1} and \ref{assumption:2} hold, we set the step-sizes as
${\eta _{\boldsymbol{x}_i}}\!=\! {\eta _{\boldsymbol{z}_i}} \! =\! \frac{2}{{L + {{\eta _{\lambda}}}M{L^2} + {{\eta _{\boldsymbol{\theta }}}}N{L^2} + 8(\frac{{M\gamma {L^2}}}{{{{\eta _{\lambda}}}{\underline{c}_1}^2}} + \frac{{N\gamma {L^2}}}{{{{\eta _{\boldsymbol{\theta }}}}{\underline{c}_2}^2}})}},\forall i$, $ {\eta _{\boldsymbol{\theta }}} \le \frac{2}{{L + 2c_2^0}} $ and   ${{\eta _{\lambda}}} \!<\! \min \{\frac{ 2}{{L + 2c_1^0}}  ,\frac{1}{{30\tau{k_1}N{L^2}}}\} $. For a given $\epsilon $, we have:
\begin{equation}
\begin{array}{l}
\!T( \epsilon ) \! \sim  \! \mathcal{O}(\max  \{ {(\frac{{4M\alpha_4}}{{{{\eta _{\lambda}}}^2}}\! +\! \frac{{4N\alpha_5}}{{{{\eta _{\boldsymbol{\theta }}}}^2}})^2}\frac{1}{{{\epsilon ^2}}}, \\ 

\qquad \quad \; \;  {(\frac{{4{{{(d_9 + \frac{{{{\eta _{\boldsymbol{\theta }}}}(N  \!-\!  S){{L}^2}}}{2})}}} (\mathop d\limits^ -  +  k_d\tau(\tau  -  1))  {d_8}}}{{{\epsilon}}} \!+\! (T_1 \!+\! 2)^{\frac{1}{2}})^2}\}), 
\end{array}
\end{equation}
where ${\alpha_4}$, ${\alpha_5}$, $\gamma$,  $k_d$, $T_1$, $M$, $N$, $S$, $\tau$, $\mathop d\limits^ -  $, ${d_8}$ and ${d_9}$ are constants. The detailed proof is given in Appendix F.  Moreover, the influence of parameters (e.g., $T_1, \tau, N, S$) in iteration complexity is discussed in Appendix F in the supplementary material.
\end{theorem}

\begin{theorem}
\label{theorem: communication complexity}
{\textbf{(Communication Complexity)}} The overall communication complexity of the proposed algorithm can be divided into complexity at every iteration and complexity of updating $\mu$-cuts, which can be expressed as $\mathcal{O}( \sum_{t=1}^{T( \epsilon )}{C_1^t} + C_2)$, where $C_1^t=32S(2\sum_{i=1}^3{d_i}+d_1 +|P_{{\rm{\uppercase\expandafter{\romannumeral2}}}}^{t}| )$, $C_2=32\sum_{t\in \mathcal{Q}} (NK(3\sum_{i=2}^3d_i+2|P_{{\rm{\uppercase\expandafter{\romannumeral2}}}}^{t}|) + N|P_{{\rm{\uppercase\expandafter{\romannumeral2}}}}^{t}|(2\sum_{i=2}^3d_i+d_1+1)  ) $, and set $\mathcal{Q}=\{ T_{{\rm{pre}}}, \cdots, \lfloor \frac{T_1}{T_{{\rm{pre}}}} \rfloor \cdot T_{{\rm{pre}}} \}$. Detailed proof is provided in Appendix J in supplementary material.

\end{theorem}

\begin{figure*}[t]
\centering
\subfigure[Diabetes]
{\begin{minipage}{4.2cm}
\label{fig:dis-diabetes}
\includegraphics[scale=0.215]{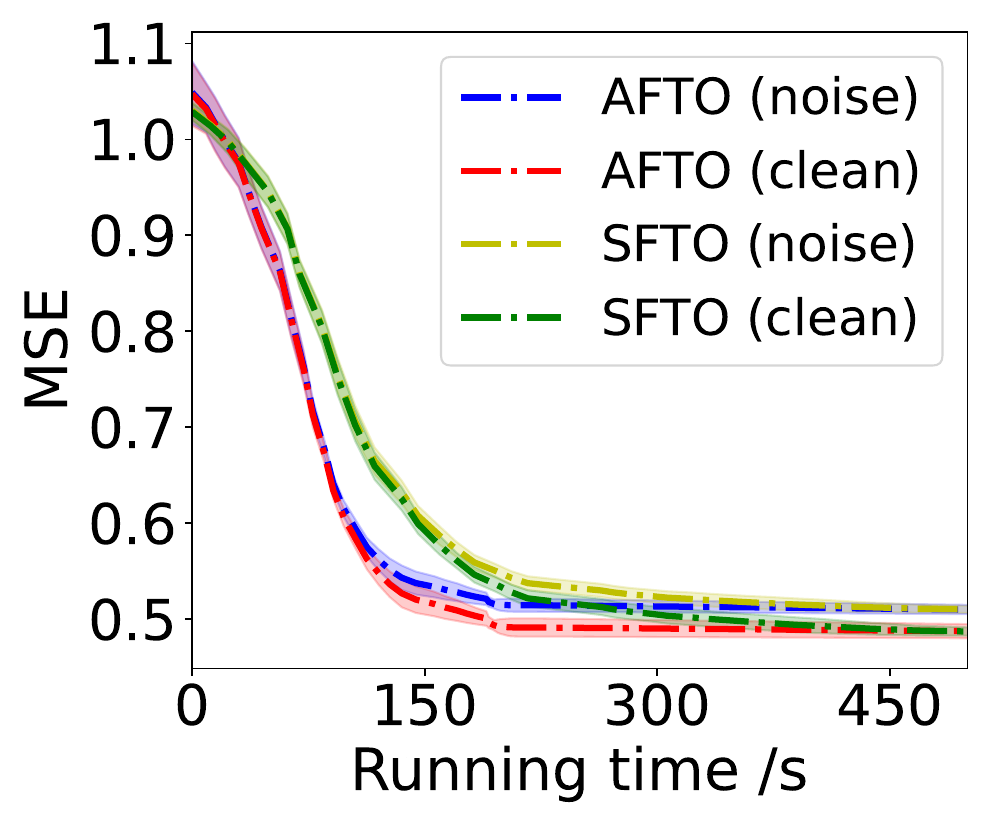}  
\end{minipage}}
\subfigure[Boston] 
{\begin{minipage}{4.2cm}
\label{fig:dis-boston}   \includegraphics[scale=0.215]{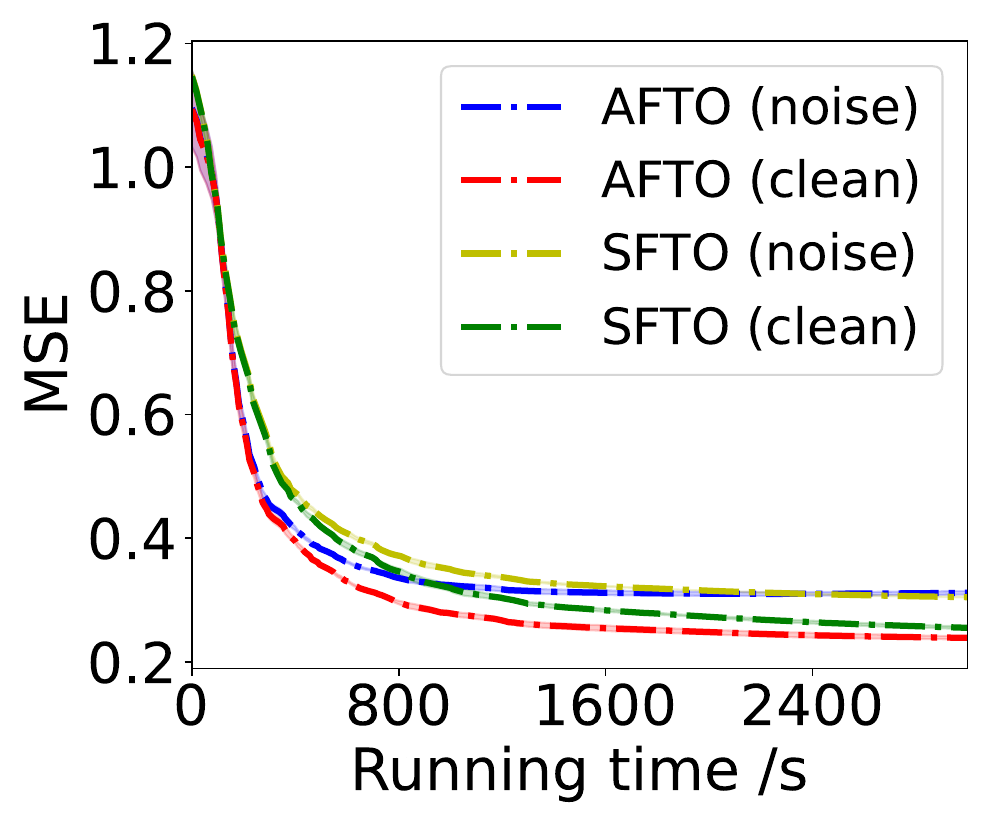}  
\end{minipage}}
\subfigure[Red-wine] 
{\begin{minipage}{4.2cm}
\label{fig:dis-red}
\includegraphics[scale=0.215]{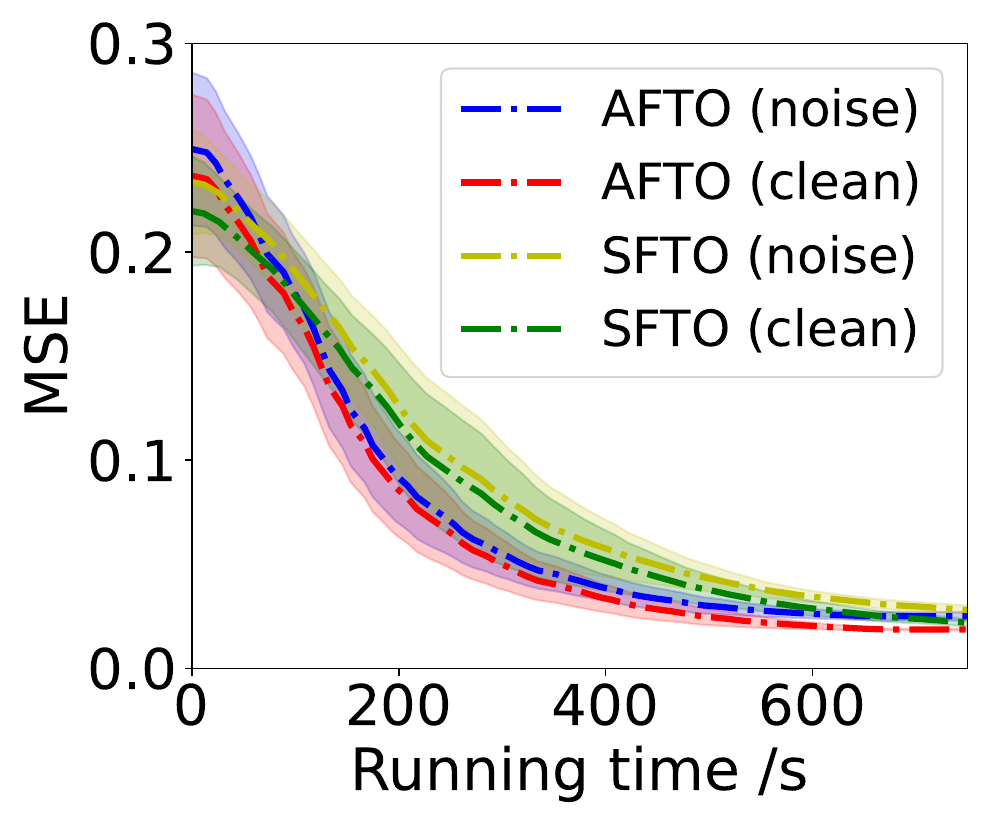}  
\end{minipage}}
\subfigure[White-wine] 
{\begin{minipage}{4.2cm}
\label{fig:dis-white}
\includegraphics[scale=0.215]{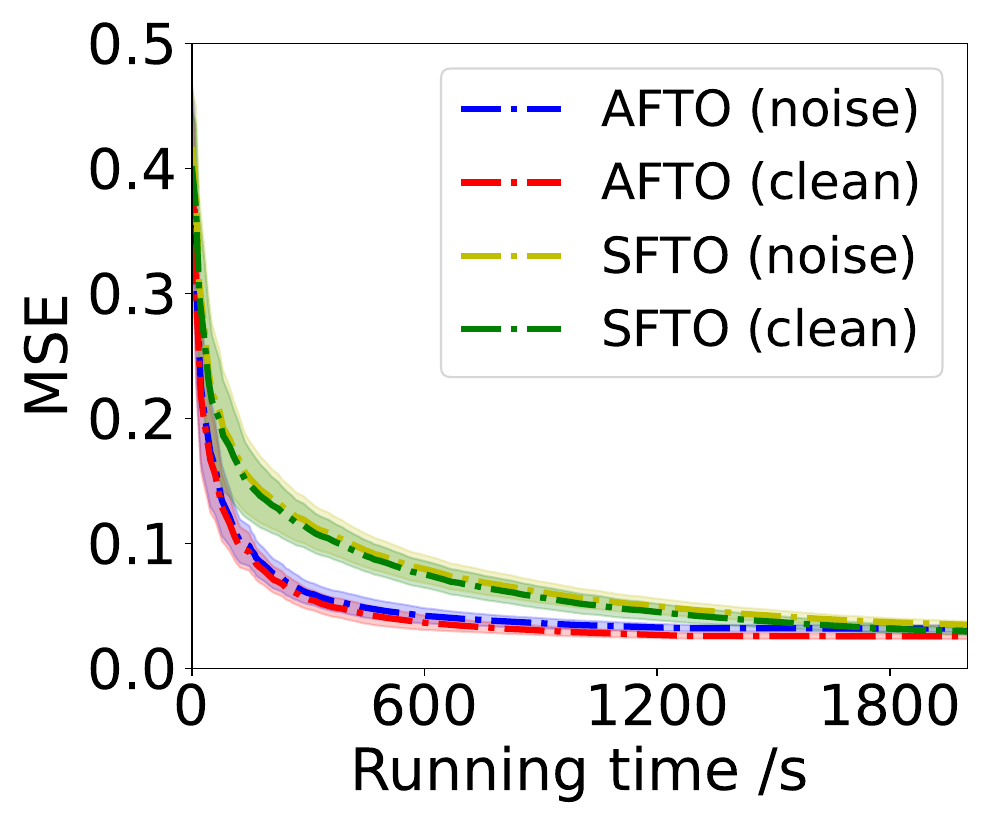}  
\end{minipage}}
\caption{MSE of clean test data and test data with Gaussian noise
on (a) Diabetes, (b) Boston, (c) Red-wine quality, and (d) White-wine quality datasets. All experiments are repeated five times, and the shaded areas represent the standard deviation.} 
\label{fig:dis-hyper}
\end{figure*}

\begin{figure*}[t]
\centering
\subfigure[Test accuracy]
{\begin{minipage}{4.2cm}
\label{fig:dis-svhn-acc}
\includegraphics[scale=0.215]{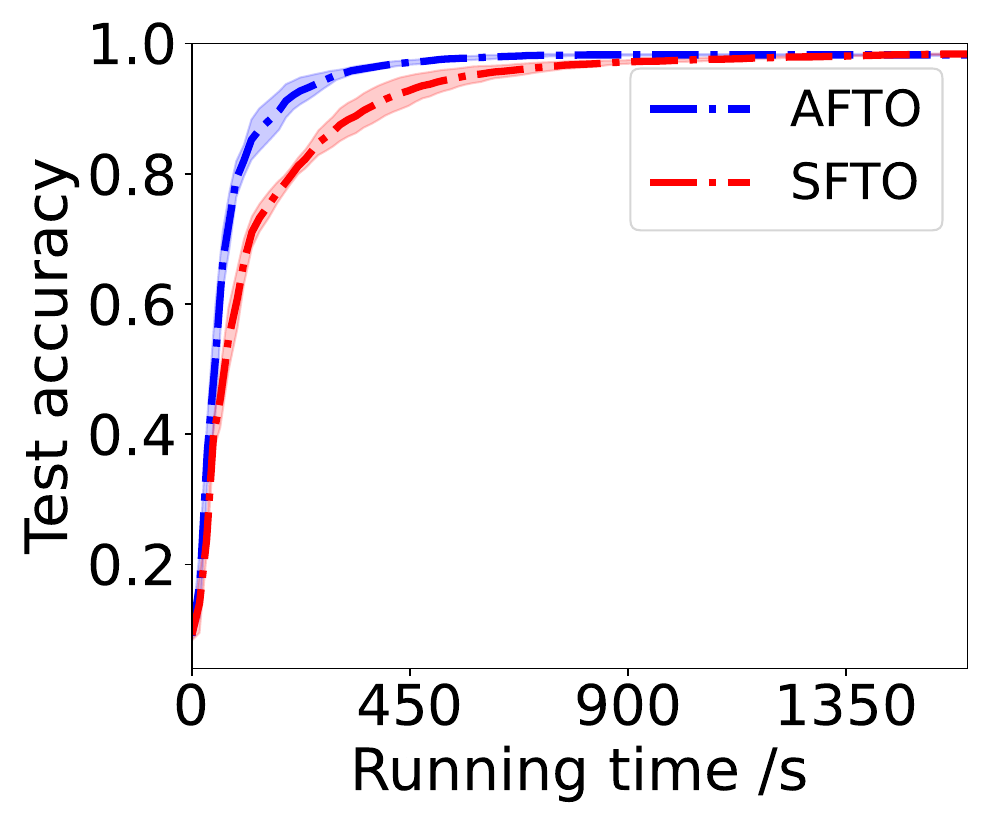}  
\end{minipage}}
\subfigure[Test loss] 
{\begin{minipage}{4.2cm}
\label{fig:dis-svhn-loss}   \includegraphics[scale=0.215]{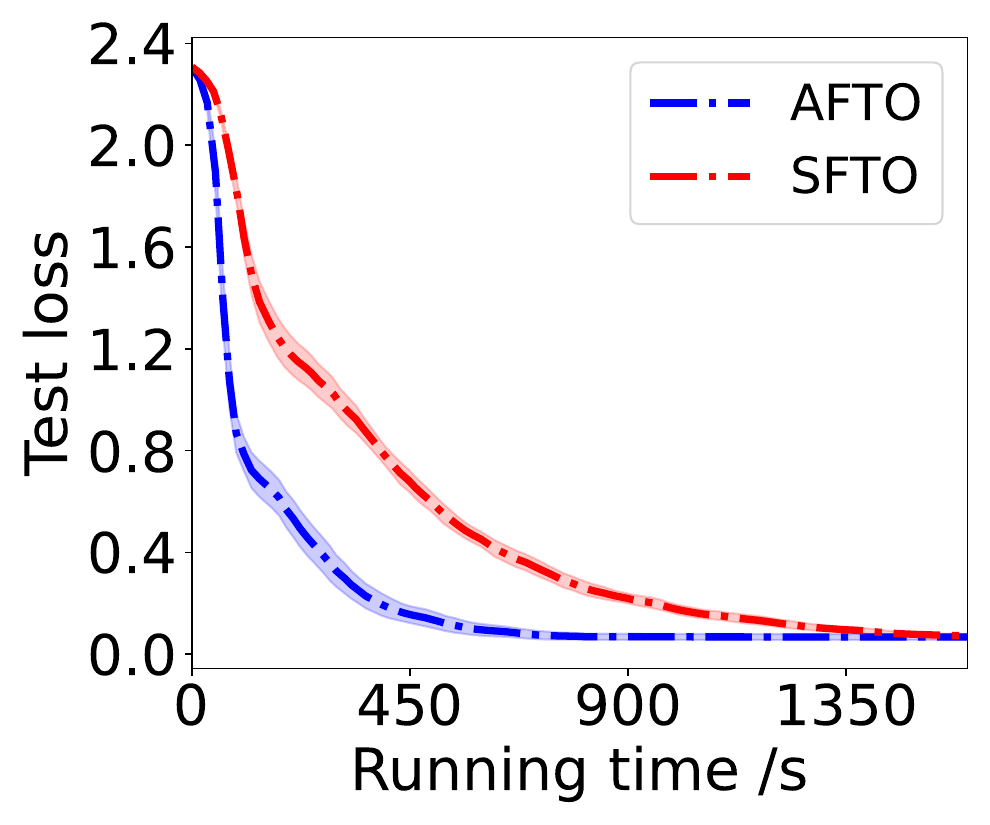}  
\end{minipage}}
\subfigure[Test accuracy] 
{\begin{minipage}{4.2cm}
\label{fig:dis-pre-mnist-acc}
\includegraphics[scale=0.215]{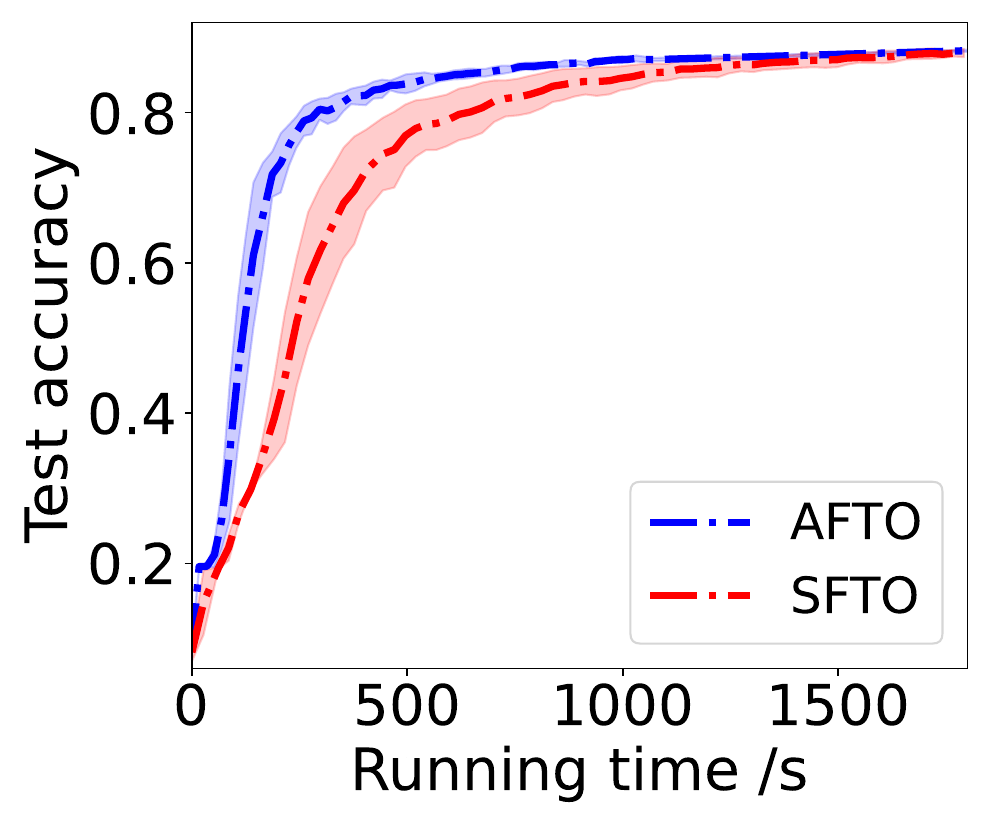}  
\end{minipage}}
\subfigure[Test loss] 
{\begin{minipage}{4.2cm}
\label{fig:dis-pre-mnist-loss}
\includegraphics[scale=0.215]{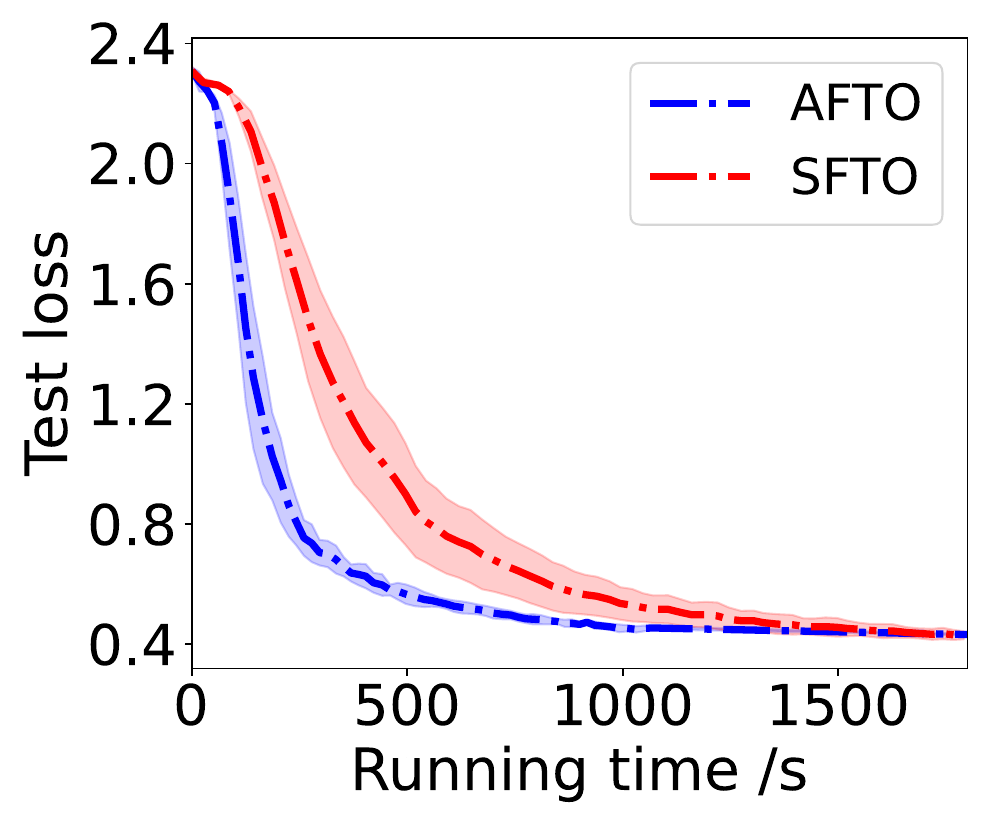}  
\end{minipage}}
\caption{(a) Test accuracy and (b) test loss vs running time when SVHN is utilized to pretrain the model. (c) Test accuracy and (d) test loss vs running time  when MNIST is utilized to pretrain the model. All experiments are repeated five times.} 
\label{fig:dis-domain}
\end{figure*}

\section{Experiment}

In the experiment, two distributed trilevel optimization tasks are employed to assess the performance of the proposed method. In the distributed robust hyperparameter optimization, experiments are carried out on the regression tasks  following \cite{sato2021gradient}, and in distributed domain adaptation for pretraining $\&$ finetuning, the multiple domain digits recognition task in \cite{qian2019robust,wang2021discriminative} is considered. The details of the experimental setting are summarized in Table \ref{tab:details_main} and Appendix H. More experimental results are reported in Appendix G. To further show the superior performance of the proposed method, experimental results of comparisons between the non-distributed version of the proposed method with existing state-of-the-art TLO methods \cite{sato2021gradient,choe2022betty} on three TLO tasks are shown in Appendix A in the supplementary material.

\subsection{Distributed Robust Hyperparameter Optimization}

The robust hyperparameter optimization \cite{sato2021gradient} aims to train a machine learning model that is robust against the noise in input data, which is inspired by bilevel hyperparameter optimization \cite{chen2022single} and adversarial training \cite{han2023fedal,zhang2022revisiting}. And we consider the following distributed robust hyperparameter optimization problem,
\begin{equation}
\begin{array}{l}
\mathop {\min }\sum\nolimits_{j} \frac{1}{|D^{\rm{val}}_j|}|| y^{\rm{val}}_j - f(X^{\rm{val}}_j;{\boldsymbol{w}})||^2 \;{\mathop{\rm s.t.}} \\

 {\boldsymbol{p}} \!=\! \mathop {\arg \max } \limits_{{\boldsymbol{p}}'} \! \sum\nolimits_{j} (\! \frac{1}{{|D^{\rm{tr}}_j|}} ||y^{\rm{tr}}_j\! -\! f(X^{\rm{tr}}_j\! +\! {p_j'} ;{\boldsymbol{w}}) ||^2 \! -\! c||{p_j'} ||^2) \,{\mathop{\rm s.t.}}\\

 {\boldsymbol{w}} \!=\! \mathop {\arg \min }\limits_{{\boldsymbol{w}}'}\!\sum\nolimits_{j}(\! \frac{1}{{|D^{\rm{tr}}_j|}} ||y^{\rm{tr}}_j\! -\! f(X^{\rm{tr}}_j \!+\! {p_j'} ;{\boldsymbol{w}}') ||^2  \!+\! e^{{\boldsymbol{\varphi }}} ||{\boldsymbol{w}}'||_{1*})  \\
{\mathop{\rm var.}} \qquad \quad {\boldsymbol{\varphi }} ,{\boldsymbol{p}},{\boldsymbol{w}},
\end{array}
\end{equation}
where ${\boldsymbol{\varphi }}$, ${\boldsymbol{w }}$ and ${\boldsymbol{p}}$ respectively denote the regularization parameter, model parameter, and adversarial noise,   ${\boldsymbol{p}}'=[p_1',\cdots,p_N']$, $N$ is the number of workers. $f$ denotes the output of a MLP, $c$ denotes the penalty for the adversarial noise, and $|| \cdot ||_{1*}$ is a  smoothed $l_1$-norm \cite{saheya2019neural}. $X^{\rm{val}}_j, y^{\rm{val}}_j, {|D^{\rm{val}}_j|}$, $X^{\rm{tr}}_j, y^{\rm{tr}}_j, {|D^{\rm{tr}}_j|}$ respectively denote the data, label and the number of data of the validation and training datasets on local worker $j$. Following \cite{sato2021gradient}, the experiments are carried out on the regression tasks with the following
datasets: Diabetes  \cite{dua2017uci}, Boston  \cite{harrison1978hedonic}, Red-wine and White-wine quality  \cite{cortez2009modeling} datasets. We summarize the experimental setting on each dataset in Table \ref{tab:details_main}. To show the performance of the proposed AFTO, we report the mean squared error (MSE) of clean test data and test data with Gaussian noise vs running time of the AFTO and SFTO (Synchronous Federated Trilevel Optimization) in Figure \ref{fig:dis-hyper}. It is seen that the proposed AFTO can effectively solve the TLO problem in a distributed manner and converges much faster than SFTO since the master can update its variables once it receives updates from a subset of workers instead of all workers in AFTO. Furthermore, we compare the proposed method with the state-of-the-art distributed bilevel optimization methods ADBO \cite{jiao2022asynchronous} and FEDNEST \cite{tarzanagh2022fednest}. It is shown in Table \ref{tab:ADBO-ATOP} that the proposed AFTO can achieve superior performance, which demonstrates the effectiveness of the proposed method.

\renewcommand\arraystretch{1.1}
\renewcommand\tabcolsep{10pt}
\begin{table}[t]
\centering
\renewcommand{\thetable}{\arabic{table}}
{
\begin{tabular}{l|c|c|c|c}
\toprule
    & $N$     & $S$ & Stragglers & $\tau$ \\ \hline
Diabetes  & 4  & 3 & 1 & 10 \\ 
Boston & 4  & 3 & 1  & 10 \\
Red-wine & 4  & 3 & 1 & 10\\
White-wine & 6  & 4 &  1 & 10 \\
SVHN (finetune) & 4  & 3 & 1  & 5 \\
SVHN (pretrain) & 6  & 3 & 2  & 15\\
\bottomrule  
\end{tabular}}
\caption{Experimental setting in distributed robust hyperparameter optimization and distributed
domain adaptation.}
\label{tab:details_main}
\end{table}

\renewcommand\arraystretch{1}
\renewcommand\tabcolsep{10pt}
\begin{table*}[t]
\centering
\renewcommand{\thetable}{\arabic{table}}
{
\begin{tabular}{l|c|c|c|c}
\toprule
 Method   & Diabetes   & Boston  & Red-wine & White-wine   \\ \hline
FEDNEST  &  0.5293 ± 0.0229 & 0.3509 ± 0.0177 &  0.0339 ± 0.0014  & 0.0268 ± 0.0010  \\
ADBO  &  0.5284 ± 0.0074 & 0.3243 ± 0.0046 & 0.0336 ± 0.0018 & 0.0277 ± 0.0013  \\ \hline
\textbf{AFTO} & \textbf{0.5124} ± \textbf{0.0068}  & \textbf{0.3130} ± \textbf{0.0037} & \textbf{0.0321} ± \textbf{0.0026} & \textbf{0.0248} ± \textbf{0.0021} \\

\bottomrule  
\end{tabular}}
\caption{MSE of test data with Gaussian noise, lower scores $\downarrow$ represent better performance which are shown in boldface.}
\label{tab:ADBO-ATOP}
\end{table*}

\subsection{Distributed Domain Adaptation}

Pretraining/finetuning paradigms are increasingly adopted recently in self-supervised
learning \cite{he2020momentum}.  In \cite{raghu2021meta}, a domain adaptation strategy is proposed, which combines data reweighting with a pretraining/finetuning framework to automatically
decrease/increase the weight of pretraining samples that cause negative/positive transfer, and can be formulated as trilevel optimization \cite{choe2022betty}. The corresponding distributed trilevel optimization problem is given as follows,
\begin{equation}
\begin{array}{l}
\; \mathop {\min } \sum\nolimits_{j} L_{FT,j}({\boldsymbol{\varphi }} ,{\boldsymbol{v}},{\boldsymbol{w}}) \;{\mathop{\rm s.t.}} \vspace{1mm}\\

\; {\boldsymbol{v}}\! =\! \mathop {\arg \min }\limits_{{\boldsymbol{v}}'} \sum\nolimits_{j} \left(L_{FT,j}({\boldsymbol{\varphi }} ,{\boldsymbol{v}}',{\boldsymbol{w}}) \!+\! \lambda ||{\boldsymbol{v}}' -{\boldsymbol{w}} ||^2  \right) \;{\mathop{\rm s.t.}} \\

\; {\boldsymbol{w}}\! =\! \mathop {\arg \min }\limits_{{\boldsymbol{w}}'} \!\sum\nolimits_{j} \! \frac{1}{\mathcal{D}_j}\!\sum\nolimits_{x_{i,j}\in\mathcal{D}_j}\! \mathcal{R}(x_{i,j}, {\boldsymbol{\varphi }})\!\cdot \! L_{PT,j}^i({\boldsymbol{\varphi }} ,{\boldsymbol{v}}',{\boldsymbol{w}}')  \\
\;  {\mathop{\rm var.}} \qquad \quad {\boldsymbol{\varphi }} ,{\boldsymbol{v}},{\boldsymbol{w}},
\end{array}
\end{equation}
where ${\boldsymbol{\varphi }}$, ${\boldsymbol{v}}$ and ${\boldsymbol{w}}$ respectively denote the parameters for pretraining, finetuning, and reweighting networks. $x_{i,j}$ and $L_{PT,j}^i$ represent the $i^{\rm{th}}$ pretraining sample and loss in worker $j$, $L_{FT,j}$ represents the finetuning loss in worker $j$. $\mathcal{R}(x_{i,j}, {\boldsymbol{\varphi }})$ denotes the importance of pretraining sample $x_{i,j}$, and $\lambda$ is the proximal regularization parameter. To evaluate the performance of the proposed method, the multiple domain digits recognition task in \cite{qian2019robust,wang2021discriminative} is considered.  There are two benchmark datasets for
this task: MNIST \cite{lecun1998gradient} and SVHN \cite{netzer2011reading}. In the experiments, we utilize the same image resize strategy as in \cite{qian2019robust} to make the format consistent, and LeNet-5 is used for all pretraining/finetuning/reweighting networks. We summarize the experimental setting in Table \ref{tab:details_main} and Appendix H. Following \cite{ji2021bilevel}, we utilize the test accuracy/test loss vs running time to evaluate the proposed AFTO. It is seen from Figure \ref{fig:dis-domain} that the proposed AFTO can effectively solve the distributed trilevel optimization problem and exhibits superior performance,  which achieves a faster convergence rate than SFTO with a maximum acceleration of approximately 80$\%$.

\section{Conclusion}

Existing trilevel learning works focus on the non-distributed setting which may lead to data privacy risks, and do not provide the non-asymptotic analysis. To this end, we propose an asynchronous federated trilevel optimization method for TLO problems. To our best knowledge, this work takes an initial step that aims to solve the TLO problems in an asynchronous federated manner. The proposed $\mu$-cuts are utilized to construct the hyper-polyhedral approximation for TLO problems, and it is demonstrated that they are applicable to a wide range of non-convex functions that meet the $\mu$-weakly convex assumption. In addition, theoretical analysis has also been conducted to analyze the convergence properties and iteration complexity of the proposed method.


\bibliography{aaai24}

\begin{thebibliography}{48}
\providecommand{\natexlab}[1]{#1}

\bibitem[{Arora and Barak(2009)}]{arora2009computational}
Arora, S.; and Barak, B. 2009.
\newblock \emph{Computational complexity: a modern approach}.
\newblock Cambridge University Press.

\bibitem[{Assran et~al.(2020)Assran, Aytekin, Feyzmahdavian, Johansson, and Rabbat}]{assran2020advances}
Assran, M.; Aytekin, A.; Feyzmahdavian, H.~R.; Johansson, M.; and Rabbat, M.~G. 2020.
\newblock Advances in asynchronous parallel and distributed optimization.
\newblock \emph{Proceedings of the IEEE}, 108(11): 2013--2031.

\bibitem[{Avraamidou(2018)}]{avraamidou2018mixed}
Avraamidou, S. 2018.
\newblock Mixed-integer multi-level optimization through multi-parametric programming.

\bibitem[{Ben-Ayed and Blair(1990)}]{ben1990computational}
Ben-Ayed, O.; and Blair, C.~E. 1990.
\newblock Computational difficulties of bilevel linear programming.
\newblock \emph{Operations Research}, 38(3): 556--560.

\bibitem[{Bertsekas(2015)}]{bertsekas2015convex}
Bertsekas, D. 2015.
\newblock \emph{Convex optimization algorithms}.
\newblock Athena Scientific.

\bibitem[{Bertsekas and Yu(2011)}]{bertsekas2011unifying}
Bertsekas, D.~P.; and Yu, H. 2011.
\newblock A unifying polyhedral approximation framework for convex optimization.
\newblock \emph{SIAM Journal on Optimization}, 21(1): 333--360.

\bibitem[{Blair(1992)}]{blair1992computational}
Blair, C. 1992.
\newblock The computational complexity of multi-level linear programs.
\newblock \emph{Annals of Operations Research}, 34.

\bibitem[{B{\"u}rger, Notarstefano, and Allg{\"o}wer(2013)}]{burger2013polyhedral}
B{\"u}rger, M.; Notarstefano, G.; and Allg{\"o}wer, F. 2013.
\newblock A polyhedral approximation framework for convex and robust distributed optimization.
\newblock \emph{IEEE Transactions on Automatic Control}, 59(2): 384--395.

\bibitem[{Chen et~al.(2022{\natexlab{a}})Chen, Feng, Guo, and Yang}]{chen2022trilevel}
Chen, S.; Feng, S.; Guo, Z.; and Yang, Z. 2022{\natexlab{a}}.
\newblock Trilevel optimization model for competitive pricing of electric vehicle charging station considering distribution locational marginal price.
\newblock \emph{IEEE Transactions on Smart Grid}, 13(6): 4716--4729.

\bibitem[{Chen et~al.(2022{\natexlab{b}})Chen, Sun, Xiao, and Yin}]{chen2022single}
Chen, T.; Sun, Y.; Xiao, Q.; and Yin, W. 2022{\natexlab{b}}.
\newblock A single-timescale method for stochastic bilevel optimization.
\newblock In \emph{International Conference on Artificial Intelligence and Statistics}, 2466--2488. PMLR.

\bibitem[{Choe et~al.(2022)Choe, Neiswanger, Xie, and Xing}]{choe2022betty}
Choe, S.~K.; Neiswanger, W.; Xie, P.; and Xing, E. 2022.
\newblock Betty: An automatic differentiation library for multilevel optimization.
\newblock \emph{arXiv preprint arXiv:2207.02849}.

\bibitem[{Cortez et~al.(2009)Cortez, Cerdeira, Almeida, Matos, and Reis}]{cortez2009modeling}
Cortez, P.; Cerdeira, A.; Almeida, F.; Matos, T.; and Reis, J. 2009.
\newblock Modeling wine preferences by data mining from physicochemical properties.
\newblock \emph{Decision support systems}, 47(4): 547--553.

\bibitem[{Davis and Drusvyatskiy(2019)}]{davis2019stochastic}
Davis, D.; and Drusvyatskiy, D. 2019.
\newblock Stochastic model-based minimization of weakly convex functions.
\newblock \emph{SIAM Journal on Optimization}, 29(1): 207--239.

\bibitem[{Dua, Graff et~al.(2017)}]{dua2017uci}
Dua, D.; Graff, C.; et~al. 2017.
\newblock UCI machine learning repository.

\bibitem[{Franc, Sonnenburg, and Werner(2011)}]{franc2011cutting}
Franc, V.; Sonnenburg, S.; and Werner, T. 2011.
\newblock Cutting plane methods in machine learning.
\newblock \emph{Optimization for Machine Learning}, 185--218.

\bibitem[{Franceschi et~al.(2018)Franceschi, Frasconi, Salzo, Grazzi, and Pontil}]{franceschi2018bilevel}
Franceschi, L.; Frasconi, P.; Salzo, S.; Grazzi, R.; and Pontil, M. 2018.
\newblock Bilevel programming for hyperparameter optimization and meta-learning.
\newblock In \emph{International Conference on Machine Learning}, 1568--1577. PMLR.

\bibitem[{Garg et~al.(2022)Garg, Zhang, Sridhara, Hosseini, Xing, and Xie}]{garg2022learning}
Garg, B.; Zhang, L.; Sridhara, P.; Hosseini, R.; Xing, E.; and Xie, P. 2022.
\newblock Learning from mistakes--a framework for neural architecture search.
\newblock In \emph{Proceedings of the AAAI Conference on Artificial Intelligence}, volume~36, 10184--10192.

\bibitem[{Gould et~al.(2016)Gould, Fernando, Cherian, Anderson, Cruz, and Guo}]{gould2016differentiating}
Gould, S.; Fernando, B.; Cherian, A.; Anderson, P.; Cruz, R.~S.; and Guo, E. 2016.
\newblock On differentiating parameterized argmin and argmax problems with application to bi-level optimization.
\newblock \emph{arXiv preprint arXiv:1607.05447}.

\bibitem[{Guo et~al.(2020)Guo, Yang, Xu, Liu, and Lin}]{guo2020meets}
Guo, M.; Yang, Y.; Xu, R.; Liu, Z.; and Lin, D. 2020.
\newblock When nas meets robustness: In search of robust architectures against adversarial attacks.
\newblock In \emph{Proceedings of the IEEE/CVF Conference on Computer Vision and Pattern Recognition}, 631--640.

\bibitem[{Han, Shi, and Huang(2023)}]{han2023fedal}
Han, P.; Shi, X.; and Huang, J. 2023.
\newblock FedAL: Black-Box Federated Knowledge Distillation Enabled by Adversarial Learning.
\newblock \emph{arXiv preprint arXiv:2311.16584}.

\bibitem[{Han, Wang, and Leung(2020)}]{han2020adaptive}
Han, P.; Wang, S.; and Leung, K.~K. 2020.
\newblock Adaptive gradient sparsification for efficient federated learning: An online learning approach.
\newblock In \emph{2020 IEEE 40th international conference on distributed computing systems (ICDCS)}, 300--310. IEEE.

\bibitem[{Harrison~Jr and Rubinfeld(1978)}]{harrison1978hedonic}
Harrison~Jr, D.; and Rubinfeld, D.~L. 1978.
\newblock Hedonic housing prices and the demand for clean air.
\newblock \emph{Journal of environmental economics and management}, 5(1): 81--102.

\bibitem[{He et~al.(2020)He, Fan, Wu, Xie, and Girshick}]{he2020momentum}
He, K.; Fan, H.; Wu, Y.; Xie, S.; and Girshick, R. 2020.
\newblock Momentum contrast for unsupervised visual representation learning.
\newblock In \emph{Proceedings of the IEEE/CVF conference on computer vision and pattern recognition}, 9729--9738.

\bibitem[{Ji, Yang, and Liang(2021)}]{ji2021bilevel}
Ji, K.; Yang, J.; and Liang, Y. 2021.
\newblock Bilevel optimization: Convergence analysis and enhanced design.
\newblock In \emph{International Conference on Machine Learning}, 4882--4892. PMLR.

\bibitem[{Jiao, Yang, and Song(2022)}]{jiao2022distributed}
Jiao, Y.; Yang, K.; and Song, D. 2022.
\newblock Distributed distributionally robust optimization with non-convex objectives.
\newblock \emph{Advances in neural information processing systems}, 35: 7987--7999.

\bibitem[{Jiao et~al.(2022{\natexlab{a}})Jiao, Yang, Song, and Tao}]{jiao2022timeautoad}
Jiao, Y.; Yang, K.; Song, D.; and Tao, D. 2022{\natexlab{a}}.
\newblock TimeAutoAD: Autonomous Anomaly Detection With Self-Supervised Contrastive Loss for Multivariate Time Series.
\newblock \emph{IEEE Transactions on Network Science and Engineering}, 9(3): 1604--1619.

\bibitem[{Jiao et~al.(2022{\natexlab{b}})Jiao, Yang, Wu, Song, and Jian}]{jiao2022asynchronous}
Jiao, Y.; Yang, K.; Wu, T.; Song, D.; and Jian, C. 2022{\natexlab{b}}.
\newblock Asynchronous Distributed Bilevel Optimization.
\newblock In \emph{The Eleventh International Conference on Learning Representations}.

\bibitem[{LeCun et~al.(1998)LeCun, Bottou, Bengio, and Haffner}]{lecun1998gradient}
LeCun, Y.; Bottou, L.; Bengio, Y.; and Haffner, P. 1998.
\newblock Gradient-based learning applied to document recognition.
\newblock \emph{Proceedings of the IEEE}, 86(11): 2278--2324.

\bibitem[{Liu et~al.(2021)Liu, Liu, Zeng, and Zhang}]{liu2021towards}
Liu, R.; Liu, Y.; Zeng, S.; and Zhang, J. 2021.
\newblock Towards gradient-based bilevel optimization with non-convex followers and beyond.
\newblock \emph{Advances in Neural Information Processing Systems}, 34: 8662--8675.

\bibitem[{Lu et~al.(2020)Lu, Tsaknakis, Hong, and Chen}]{lu2020hybrid}
Lu, S.; Tsaknakis, I.; Hong, M.; and Chen, Y. 2020.
\newblock Hybrid block successive approximation for one-sided non-convex min-max problems: algorithms and applications.
\newblock \emph{IEEE Transactions on Signal Processing}, 68: 3676--3691.

\bibitem[{Netzer et~al.(2011)Netzer, Wang, Coates, Bissacco, Wu, and Ng}]{netzer2011reading}
Netzer, Y.; Wang, T.; Coates, A.; Bissacco, A.; Wu, B.; and Ng, A.~Y. 2011.
\newblock Reading digits in natural images with unsupervised feature learning.

\bibitem[{Qian et~al.(2019)Qian, Zhu, Tang, Jin, Sun, and Li}]{qian2019robust}
Qian, Q.; Zhu, S.; Tang, J.; Jin, R.; Sun, B.; and Li, H. 2019.
\newblock Robust optimization over multiple domains.
\newblock In \emph{Proceedings of the AAAI Conference on Artificial Intelligence}, volume~33, 4739--4746.

\bibitem[{Raghu et~al.(2021)Raghu, Lorraine, Kornblith, McDermott, and Duvenaud}]{raghu2021meta}
Raghu, A.; Lorraine, J.; Kornblith, S.; McDermott, M.; and Duvenaud, D.~K. 2021.
\newblock Meta-learning to improve pre-training.
\newblock \emph{Advances in Neural Information Processing Systems}, 34: 23231--23244.

\bibitem[{Saheya, Nguyen, and Chen(2019)}]{saheya2019neural}
Saheya, B.; Nguyen, C.~T.; and Chen, J.-S. 2019.
\newblock Neural network based on systematically generated smoothing functions for absolute value equation.
\newblock \emph{Journal of Applied Mathematics and Computing}, 61(1): 533--558.

\bibitem[{Sato, Tanaka, and Takeda(2021)}]{sato2021gradient}
Sato, R.; Tanaka, M.; and Takeda, A. 2021.
\newblock A Gradient Method for Multilevel Optimization.
\newblock \emph{Advances in Neural Information Processing Systems}, 34: 7522--7533.

\bibitem[{Sinha, Malo, and Deb(2017)}]{sinha2017review}
Sinha, A.; Malo, P.; and Deb, K. 2017.
\newblock A review on bilevel optimization: From classical to evolutionary approaches and applications.
\newblock \emph{IEEE Transactions on Evolutionary Computation}, 22(2): 276--295.

\bibitem[{Su et~al.(2022)Su, Zhang, Cai, Ren, Wang, Yi, Song, Chen, Deng, Xu et~al.}]{su2022gba}
Su, W.; Zhang, Y.; Cai, Y.; Ren, K.; Wang, P.; Yi, H.; Song, Y.; Chen, J.; Deng, H.; Xu, J.; et~al. 2022.
\newblock GBA: A Tuning-free Approach to Switch between Synchronous and Asynchronous Training for Recommendation Models.
\newblock \emph{Advances in Neural Information Processing Systems}, 35: 29525--29537.

\bibitem[{Subramanya and Riggio(2021)}]{subramanya2021centralized}
Subramanya, T.; and Riggio, R. 2021.
\newblock Centralized and federated learning for predictive {VNF} autoscaling in multi-domain {5G} networks and beyond.
\newblock \emph{IEEE Transactions on Network and Service Management}, 18(1): 63--78.

\bibitem[{Tarzanagh et~al.(2022)Tarzanagh, Li, Thrampoulidis, and Oymak}]{tarzanagh2022fednest}
Tarzanagh, D.~A.; Li, M.; Thrampoulidis, C.; and Oymak, S. 2022.
\newblock Fednest: Federated bilevel, minimax, and compositional optimization.
\newblock In \emph{International Conference on Machine Learning}, 21146--21179. PMLR.

\bibitem[{Tawarmalani and Sahinidis(2005)}]{tawarmalani2005polyhedral}
Tawarmalani, M.; and Sahinidis, N.~V. 2005.
\newblock A polyhedral branch-and-cut approach to global optimization.
\newblock \emph{Mathematical programming}, 103(2): 225--249.

\bibitem[{Trombettoni et~al.(2011)Trombettoni, Araya, Neveu, and Chabert}]{trombettoni2011inner}
Trombettoni, G.; Araya, I.; Neveu, B.; and Chabert, G. 2011.
\newblock Inner regions and interval linearizations for global optimization.
\newblock In \emph{Proceedings of the AAAI Conference on Artificial Intelligence}, volume~25, 99--104.

\bibitem[{Wang et~al.(2021)Wang, Chen, Lin, Sigal, and de~Silva}]{wang2021discriminative}
Wang, J.; Chen, J.; Lin, J.; Sigal, L.; and de~Silva, C.~W. 2021.
\newblock Discriminative feature alignment: Improving transferability of unsupervised domain adaptation by Gaussian-guided latent alignment.
\newblock \emph{Pattern Recognition}, 116: 107943.

\bibitem[{Xie, Koyejo, and Gupta(2019)}]{xie2019asynchronous}
Xie, C.; Koyejo, S.; and Gupta, I. 2019.
\newblock Asynchronous federated optimization.
\newblock \emph{arXiv preprint arXiv:1903.03934}.

\bibitem[{Xu et~al.(2020)Xu, Zhang, Xu, and Lan}]{xu2020unified}
Xu, Z.; Zhang, H.; Xu, Y.; and Lan, G. 2020.
\newblock A unified single-loop alternating gradient projection algorithm for nonconvex-concave and convex-nonconcave minimax problems.
\newblock \emph{arXiv preprint arXiv:2006.02032}.

\bibitem[{Yang et~al.(2014)Yang, Huang, Wu, Wang, and Chiang}]{yang2014distributed}
Yang, K.; Huang, J.; Wu, Y.; Wang, X.; and Chiang, M. 2014.
\newblock Distributed robust optimization ({DRO}), part {I}: {Framework} and example.
\newblock \emph{Optimization and Engineering}, 15(1): 35--67.

\bibitem[{Yang et~al.(2008)Yang, Wu, Huang, Wang, and Verd{\'u}}]{yang2008distributed}
Yang, K.; Wu, Y.; Huang, J.; Wang, X.; and Verd{\'u}, S. 2008.
\newblock Distributed robust optimization for communication networks.
\newblock In \emph{IEEE INFOCOM 2008-The 27th Conference on Computer Communications}, 1157--1165. IEEE.

\bibitem[{Zhang and Kwok(2014)}]{zhang2014asynchronous}
Zhang, R.; and Kwok, J. 2014.
\newblock Asynchronous distributed {ADMM} for consensus optimization.
\newblock In \emph{International conference on machine learning}, 1701--1709. PMLR.

\bibitem[{Zhang et~al.(2022)Zhang, Zhang, Khanduri, Hong, Chang, and Liu}]{zhang2022revisiting}
Zhang, Y.; Zhang, G.; Khanduri, P.; Hong, M.; Chang, S.; and Liu, S. 2022.
\newblock Revisiting and advancing fast adversarial training through the lens of bi-level optimization.
\newblock In \emph{International Conference on Machine Learning}, 26693--26712. PMLR.

\end{thebibliography}

\end{document}